\documentclass[a4paper,fleqn]{cas-dc}

\usepackage[authoryear]{natbib}
\usepackage{tabularray}
\usepackage[table]{xcolor}

\usepackage{algorithm}
\usepackage{algorithmic}

\def\tsc#1{\csdef{#1}{\textsc{\lowercase{#1}}\xspace}}
\tsc{WGM}
\tsc{QE}
\tsc{EP}
\tsc{PMS}
\tsc{BEC}
\tsc{DE}

\begin{document}
\let\WriteBookmarks\relax
\let\printorcid\relax
\def\floatpagepagefraction{1}
\def\textpagefraction{.001}
\shorttitle{Test-Time Generative Augmentation}
\shortauthors{Ma et~al.}

\title [mode = title]{Test-time generative augmentation for
 medical image segmentation}                      

\author[1,2]{Xiao Ma}[style=chinese]

\author[3,4]{Yuhui Tao}[style=chinese]

\author[1]{Zetian Zhang}[style=chinese]

\author[5]{Yuhan Zhang}[style=chinese]

\author[6,7]{Xi Wang}[style=chinese]

\author[2,8]{Sheng Zhang}[style=chinese]

\author[1]{Zexuan Ji}[style=chinese]

\author[1]{Yizhe Zhang}[style=chinese]
\cormark[1]
\ead{yizhe.zhang.cs@gmail.com}

\author[1]{Qiang Chen}[style=chinese]
\cormark[1]
\ead{chen2qiang@njust.edu.cn}

\author[2,8,9,10]{Guang Yang}[style=chinese]

\affiliation[1]{organization={School of Computer Science and Engineering, Nanjing University of Science and Technology},
                city={Nanjing},
                country={China}}
\affiliation[2]{organization={Bioengineering Department and Imperial-X, Imperial College London},
                city={London},
                postcode={W12 7SL}, 
                country={UK}}
\affiliation[3]{organization={Digital Medical Research Center, School of Basic Medical Sciences, Fudan University},
                city={Shanghai},
                country={China}}
\affiliation[4]{organization={Shanghai Key Laboratory of MICCAI},
                city={Shanghai},
                country={China}}
\affiliation[5]{organization={School of Biomedical Engineering, Shenzhen University},
                city={Shenzhen},
                country={China}}
\affiliation[6]{organization={Department of Computer Science and Engineering, The Hong Kong University of Science and Technology},
                city={Hong Kong},
                country={China}}
\affiliation[7]{organization={Department of Computer Science and Engineering, The Chinese University of Hong Kong},
                city={Hong Kong},
                country={China}}
\affiliation[8]{organization={National Heart and Lung Institute, Imperial College London},
                city={London},
                postcode={SW7 2AZ}, 
                country={UK}}
\affiliation[9]{organization={Cardiovascular Research Centre, Royal Brompton Hospital},
                city={London},
                postcode={SW3 6NP}, 
                country={UK}}
\affiliation[10]{organization={School of Biomedical Engineering \& Imaging Sciences, King's College London},
                city={London},
                postcode={WC2R 2LS}, 
                country={UK}}
                
\cortext[cor1]{Corresponding authors}

\begin{abstract}
Medical image segmentation is critical for clinical diagnosis, treatment planning, and monitoring, yet segmentation models often struggle with uncertainties stemming from occlusions, ambiguous boundaries, and variations in imaging devices. Traditional test-time augmentation (TTA) techniques typically rely on predefined geometric and photometric transformations, limiting their adaptability and effectiveness in complex medical scenarios. In this study, we introduced \textbf{T}est-\textbf{T}ime \textbf{G}enerative \textbf{A}ugmentation (\textbf{TTGA}), a novel augmentation strategy specifically tailored for medical image segmentation at inference time. Different from conventional augmentation strategies that suffer from excessive randomness or limited flexibility, TTGA leverages a domain-fine-tuned generative model to produce contextually relevant and diverse augmentations tailored to the characteristics of each test image. Built upon diffusion model inversion, a masked null-text inversion method is proposed to enable region-specific augmentations during sampling. Furthermore, a dual denoising pathway is designed to balance precise identity preservation with controlled variability. We demonstrate the efficacy of our TTGA through extensive experiments across three distinct segmentation tasks spanning nine datasets. Our results consistently demonstrate that TTGA not only improves segmentation accuracy (with DSC gains ranging from 0.1\% to 2.3\% over the baseline) but also offers pixel-wise error estimation (with DSC gains ranging from 1.1\% to 29.0\% over the baseline). The source code and demonstration are available at: \url{https://github.com/maxiao0234/TTGA}.

\end{abstract}




\maketitle

\let\thefootnote\relax\footnotetext{
\noindent \textit{© 2025. This manuscript version is made available under the CC-BY-NC-ND 4.0 license (\url{http://creativecommons.org/licenses/by-nc-nd/4.0/}).} \\
\noindent \textit{Accepted for publication in Medical Image Analysis (MedIA), Dec 2025.} \\
\noindent \textit{Official publication: March 2026. Finalized version available at: \url{https://doi.org/10.1016/j.media.2025.103902}}
}

\section{Introduction}

\begin{figure*}
\centering
\includegraphics[width=1.0\textwidth]{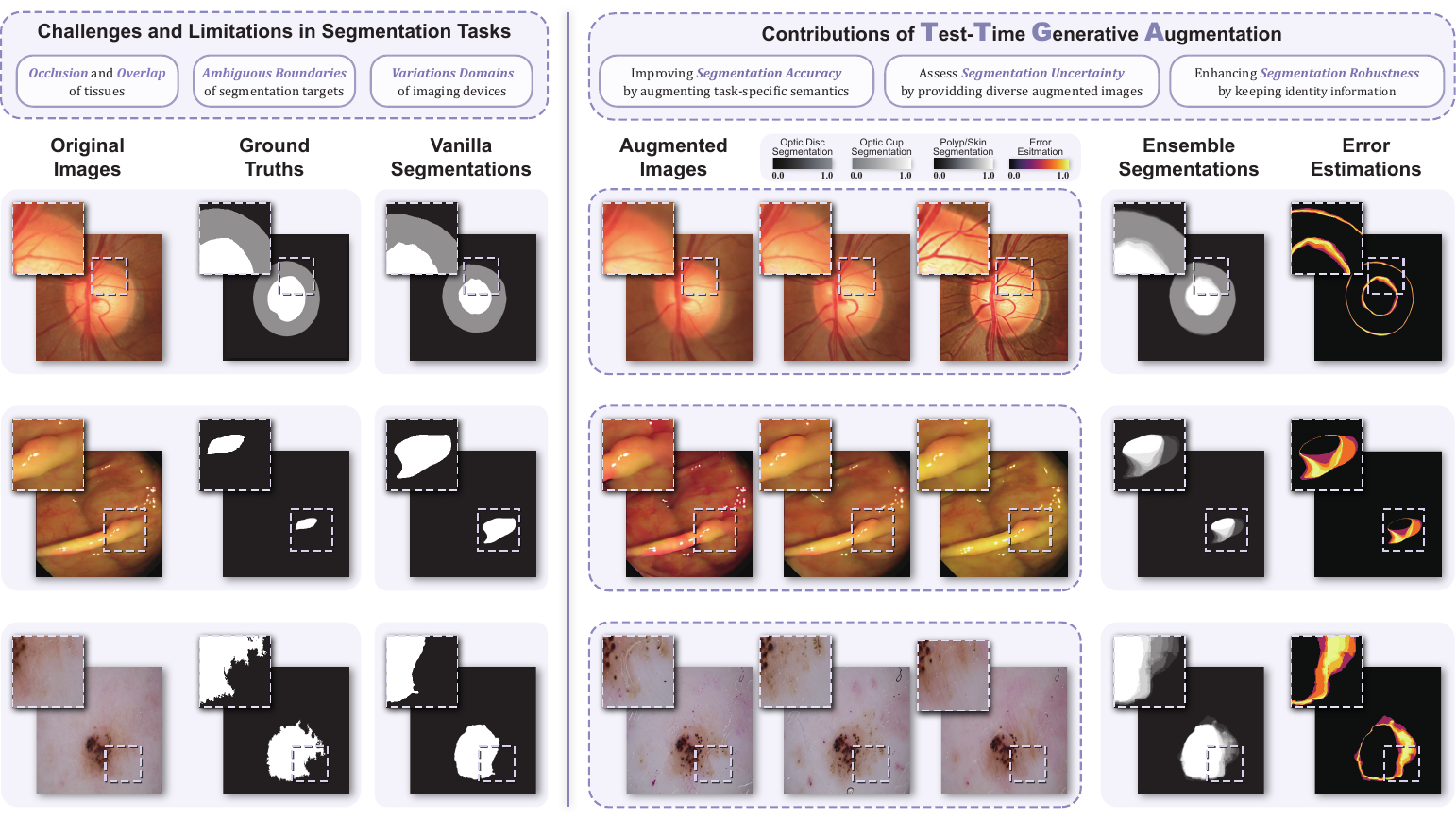}
\caption{Visualization of TTGA augmentation results on three exemplary images. The original images present challenges for segmentation due to tissue overlap, blurred boundaries, and diverse acquisition conditions. TTGA-augmented images introduce variations in local structure, sharpness, and imaging style, which enhance segmentation accuracy and robustness, while also supporting uncertainty estimation and model reliability. The color bars indicate the scales for segmentation probability and error estimation, respectively}
\label{fig-1}
\end{figure*}

Medical image segmentation plays a pivotal role in clinical decision-making, enabling precise delineation of anatomical and pathological structures for diagnosis, treatment planning, and monitoring~\cite{roy2023mednext,ozbey2023unsupervised,bateson2022source}. Despite considerable advancements in segmentation model architectures~\cite{ronneberger2015u,lin2022ds}, training paradigms~\cite{xu2019camel,ma2020ms,xu2019camel}, and data usage~\cite{qin2021efficient}, segmentation accuracy remains vulnerable to inherent uncertainties and variability in medical imaging (as depicted in Fig.~\ref{fig-1}), such as interference from occluded or overlapping non-target tissues, ambiguous boundaries, and variations in imaging devices and acquisition protocols. These challenges not only reduce segmentation accuracy but also affect the confidence and robustness of segmentation results, highlighting the necessity for improved uncertainty estimation and enhanced robustness.

During inference, partial access to the test‐time sample distribution affords a valuable opportunity to further refine state‐of‐the‐art segmentation models, and contemporary approaches may be grouped into three primary paradigms. Test‐Time Model Adaptation (TTMA) methods (e.g., ~\cite{zhang2024testfit}) iteratively update model parameters using newly observed test samples, thereby facilitating on‐the‐fly adaptation, yet they inherently gravitate toward a single optimal parameter configuration and thus cannot reliably produce the diverse ensemble of predictions needed to quantify segmentation uncertainty. In contrast, Test‐Time Dropout (TTD) techniques (e.g., ~\cite{gal2016dropout}) introduce stochastic perturbations via random neuron deactivation during inference to sample from the model’s posterior, offering valuable insights into predictive uncertainty; however, such uncontrolled randomness may undermine segmentation accuracy and yield unstable outputs. Finally, Test‐Time Augmentation (TTA) strategies (e.g., ~\cite{ayhan2022test}) generate multiple segmentations by applying a fixed set of geometric and photometric transforms to each test image and subsequently aggregating the results, thereby enhancing robustness through ensembling; nevertheless, the reliance on a predetermined transformation suite limits their adaptability to the complex, heterogeneous characteristics of medical imaging data.

To address these limitations, we introduce \textbf{Test-time Generative Augmentation (TTGA)}, a novel generative augmentation strategy tailored specifically for medical image segmentation tasks. TTGA utilizes a domain-fine-tuned generative model to generate diverse, contextually relevant augmentations of test samples during inference. TTGA utilizes a fine-tuned diffusion model to generate augmentations tailored to the specific content and semantics of each test image, offering distinct advantages over traditional methods. (1) By preserving global layout and enriching local structure distributions, TTGA mitigates interference caused by occlusion and overlap, thereby significantly improving segmentation accuracy. (2) To tackle ambiguous boundaries, TTGA introduces augmentation strategies that sharpen boundary details and simulate realistic noise variations, providing effective data perturbations that facilitate reliable uncertainty estimation. (3) To enhance robustness against variability arising from diverse imaging devices and acquisition conditions, TTGA incorporates task-specific semantics and maintains essential identity information, generating realistic variations that improve model stability across different imaging contexts.

At the core of TTGA is a novel masked null-text inversion technique, which enables region-specific augmentation by balancing identity preservation and controlled variability through dual denoising paths. This ensures critical regions remain intact while allowing meaningful variations elsewhere. This region-specific approach reflects the intuition that different parts of an image contribute unequally to segmentation decisions, enabling perturbations that reveal uncertainties and improve robustness.

Using the augmented samples, we generate multiple segmentation maps with the pre-trained model and aggregate them to produce a refined segmentation output that surpasses the quality of the original prediction. Additionally, the diversity of these maps facilitates the construction of a pixel-wise error estimation map, highlighting areas prone to segmentation inaccuracies. Our contributions are three aspects:

\begin{itemize}
\item We introduce test-time generative augmentation (TTGA), a method that integrates test sample content with a generative model fine-tuned on medical data, producing realistic, locally randomized augmentations that enhance segmentation performance.

\item We develop a novel masked null-text inversion technique for diffusion-based image editing, applying distinct conditioning to different regions during denoising to balance preservation and augmentation, thereby generating samples optimized for segmentation tasks.

\item We validate TTGA through extensive experiments across three medical image segmentation tasks, demonstrating its effectiveness in improving segmentation accuracy and providing reliable error estimation.
\end{itemize}

\section{Related work}
\subsection{Test-time augmentation}
Our method is a novel type of test-time augmentation (TTA) method. In this section, we give an overview of the existing TTA methods, highlighting the advantages and novel aspects of the proposed method against them. The process of TTA typically involves applying transformations such as flipping, rotation, scaling, cropping, or color jittering to the test images before feeding them into the model for inference. By considering multiple views of the same test instance, TTA generates better segmentation output by ensemble or fusing segmentation from the multi-views~\cite{wang2019automatic,moshkov2020test}. In addition, the variances among the segmentation of the views can be used for pixel-wise error estimation~\cite{wang2019aleatoric,ayhan2022test}. Existing TTA methods either rely on generic geometric transforms to create multiple views or rely on designing specific transforms/functions for particular segmentation tasks, which affords only a limited augmentation space and constrains the ability to effect meaningful alterations of image content. Our proposed generative augmentation (TTGA) is generally applicable to a wide range of medical image segmentation tasks and is agnostic to the segmentation model used. The proposed TTGA, via utilizing a state-of-the-art generative model, can create new views of a test sample based on the image content and data distribution, superior to hand-crafted transforms and functions.

\subsection{Generative methods based on diffusion Models}
Diffusion Models have transformed the landscape of image synthesis, offering the capability to produce high-fidelity images that capture complex data distributions. In the realm of medical imaging, these models are instrumental for generating synthetic data, enhancing image quality, and enabling advanced image editing tasks. Diffusion-based models, such as denoising diffusion probabilistic models (DDPMs) \cite{ho2020denoising} and latent diffusion models (LDMs) \cite{rombach2022high}, have become prominent for their ability to generate realistic images from noise, with applications extending to medical domains. These models support conditional generation, allowing for precise control over image content through techniques such as classifier-free guidance \cite{ho2022classifier}. Recent advancements have seen the development of physics-inspired generative models tailored for medical imaging \cite{hein2024physics}. These models have been applied to tasks such as denoising PET images, demonstrating strong generalizability across different low-dose levels, scanners, and protocols. In the context of TTGA, we leverage these advancements by fine-tuning state-of-the-art generative models with domain-specific data to ensure that the augmented images are both medically plausible and semantically consistent with the original test samples. This approach not only enhances the robustness of segmentation models but also provides a framework for more effective test-time augmentation tailored to the unique characteristics of medical images, building on the foundation laid by recent innovations in generative modeling.

\subsection{Generative methods for image editing}
Generative models have garnered significant attention owing to their exceptional realism and diversity \cite{ho2020denoising,rombach2022high,zhang2023adding}. As a form of image-to-image translation, image editing enables targeted modifications to specific content within an image while maintaining consistency with the remaining content \cite{huang2024diffusion}. In a systematic review \cite{oulmalme2025systematic}, GANs, transformers, and diffusion models were compared for medical image enhancement, emphasizing their capabilities in improving image quality and diagnostic accuracy. This review underscores the potential of generative models to enhance medical images, which is critical for tasks like segmentation and anomaly detection. However, limited by the lack of accompanying textual data and the difficulties associated with employing text-driven control methods, region-based control emerges as a more feasible approach for medical data. In this study, we devised a novel image augmentation strategy tailored for medical images through diffusion inversion. Specifically, we build upon these developments by introducing masked null-text inversion, which enables region-specific augmentations by applying different guidance conditions during the denoising process. This approach ensures that identity-preserving regions remain largely unaltered while augmentation-enhancing regions introduce necessary variability, thereby enhancing the robustness of segmentation models without compromising the diagnostic utility of the images. This method aligns with recent trends in generative AI for medical imaging, offering a novel perspective on test-time augmentation through advanced image editing techniques.

\section{Preliminaries}
To facilitate a clear understanding of our proposed approach, we first introduce foundational concepts and methodologies relevant to generative models, specifically diffusion-based models and classifier-free guidance, which form the basis of our generative augmentation strategy.
\subsection{Latent diffusion models}
Diffusion probabilistic models (DPMs) have recently gained widespread attention in generative modeling due to their ability to produce high-quality, diverse samples. Denoising diffusion probabilistic models (DDPMs) \cite{ho2020denoising} progressively remove noise from a random Gaussian vector through a Markovian denoising process in the image space, eventually arriving at a clean image sample. Latent diffusion models (LDMs) \cite{rombach2022high} adapt this approach by performing the diffusion process in a learned latent space rather than the raw image space. More concretely, given an image $I_0 \in \mathbb{R}^{H \times W \times 3}$, a pretrained variational autoencoder (VAE) is used to encode it into a latent representation $x_0=\mathcal{E}(I_0)$ and decode it back via $I_0=\mathcal{D}(x_0)$, where $\mathcal{E}$ and $\mathcal{D}$ are the encoder and decoder of the VAE, respectively, and $x_0 \in \mathbb{R}^{h \times w \times d}$ is the latent vector corresponding to the image. By operating in the latent space, LDMs can substantially reduce the computational cost of the diffusion process while still preserving essential features necessary for generating high-fidelity images.

DPMs define a forward diffusion process $\left\{x_t\right\}_{t=0}^T$ where noise is gradually added over $T$ steps, eventually producing a noisy sample $x_T \sim \mathcal{N}(\mathbf{0}, \mathbf{I})$. This process is expressed as:  
\begin{equation}
\begin{split}
    q(x_{1:T}\mid x_0) &:= \prod_{t=1}^{T} q(x_t\mid x_{t-1}), \\
    \text{where} \; q(x_t\mid x_{t-1}) &:= \mathcal{N}\left(x_t; \sqrt{\alpha_t} x_{t-1}, (1-\alpha_t) \mathbf{I} \right),
\end{split}
\label{forward_markovian}
\end{equation}
where $\left\{\alpha_t\right\}_{t=1}^T$ is a predefined schedule. From this definition, a closed-form solution can be derived: 
\begin{equation} 
q(x_t\mid x_0) = \mathcal{N} \left(x_t; \sqrt{\bar{\alpha}_t}x_0, \sqrt{1-\bar{\alpha}_t}\mathbf{I} \right),
\label{forward_non_markovian} 
\end{equation} 
where $\bar{\alpha}_t := \prod_{i=1}^t \alpha_i$. It allows directly sampling $x_t$ from $x_0$ without strictly following each intermediate step. This provides flexibility for non-Markovian noise addition strategies.

Given a trained LDM, the reverse diffusion process starts from a noisy sample $x_T$ and iteratively refines it to approximate a sample $x_0$ from the data distribution. The reverse step is given by: 
\begin{equation} p_\theta(x_{t-1}\mid x_t) := \mathcal{N} \left(x_{t-1}; \mu_\theta(x_t,t), \sigma_t \mathbf{I} \right),
\end{equation} 
where $\sigma_t$ is a known time-dependent variance term and the mean is determined by a neural network $\epsilon_\theta(x_t,t)$ trained to predict the noise: 
\begin{equation} \mu_\theta(x_t, t) = \frac{1}{\sqrt{\alpha_t}}\left(x_t - \frac{1-\bar{\alpha}_t}{\sqrt{1-\bar{\alpha}_t}}\epsilon_\theta(x_t, t)\right).
\end{equation} 
By focusing on a latent representation, LDMs achieve a favorable balance between computational efficiency and the fidelity of generated samples.

\subsection{Classifier-free guidance}
In DPMs, conditions can be incorporated into the generation process by training on image-condition pairs. Within a stochastic differential equation (SDE) framework \cite{dhariwal2021diffusion,song2020score}, the gradient of the log-probability of a conditional distribution can be decomposed as: \begin{equation} 
\nabla_{x_t} \log p(x_t\mid c) \approx \nabla_{x_t} \log p(x_t) + \omega \nabla_{x_t} \log p(c\mid x_t),
\label{sde-single} 
\end{equation} 
where $c$ is the conditioning input, and $\omega$ is a guidance scale that controls how strongly the condition influences the generation. In the SDE-based framework, score matching techniques are used to estimate $\nabla_{x_t} \log p(x_t)$ by relating it to the model’s predicted noise $\epsilon_\theta (x_t, t)$. During inference, the model leverages the learned relationship between predicted noise and the data gradient to guide the sampling process.

Classifier-free guidance \cite{ho2022classifier} achieves conditional generation without a separately trained classifier. Instead, the model is trained with both conditional and unconditional inputs. The noise prediction under classifier-free guidance is: 
\begin{equation}
\dot{\epsilon}_\theta(x_t,t,\varnothing,c) = \epsilon_\theta(x_t,t,\varnothing) + \omega \bigl(\epsilon_\theta(x_t,t,c) - \epsilon_\theta(x_t,t,\varnothing)\bigr),
\label{classifier-free-single} 
\end{equation} 
where $\varnothing$ denotes the unconditional embedding, $\epsilon_\theta(x_t, t, \varnothing)$ denotes the unconditional prediction, and $\epsilon_\theta(x_t, t, c)$ represents the conditional prediction. By adjusting $\omega$, one can flexibly tune the trade-off between diversity and adherence to the condition, making classifier-free guidance a widely adopted technique in conditional generative modeling.

\subsection{DDIM inversion}
Denoising diffusion implicit models (DDIMs) \cite{song2020denoising} generalize the DDPM reverse diffusion steps into a class of deterministic non-Markovian sampling procedures, offering improved sampling speed and quality. The relationship between consecutive steps in DDIM sampling can be written as: 
\begin{equation} 
\frac{x_{t-1}}{\sqrt{\bar{\alpha}_{t-1}}} = \frac{x_t}{\sqrt{\bar{\alpha}_t}} + \left(\sqrt{\frac{1-\bar{\alpha}_{t-1}}{\bar{\alpha}_{t-1}}} - \sqrt{\frac{1-\bar{\alpha}_t}{\bar{\alpha}_t}}\right)\epsilon_\theta(x_t, t).
\label{ddim_sample} 
\end{equation} 

This formulation can be extended to arbitrary time intervals $\Delta t$ using an ordinary differential equation (ODE) interpretation: 
\begin{equation} 
\bar{x}_{t - \Delta t} = \bar{x}_{t} + \left( \gamma_{t - \Delta t} - \gamma_{t} \right) \epsilon_\theta(x_t, t),
\label{ddim_sample_delta} 
\end{equation} 
where $\bar{x}_i = x_i/\sqrt{\bar{\alpha}_i}$ and $\gamma_i = \sqrt{(1-\bar{\alpha}_i)/\bar{\alpha}_i}$. This generalized formulation makes the DDIM sampling more adaptable by changing the step size based on the desired resolution or efficiency of the denoising process. 

To facilitate inversion within the DDIM framework, approximations are introduced to reconstruct intermediate states from generated samples. Specifically, we use $\epsilon_\theta(x_{t-1}, t-1)$ to approximate $\epsilon_\theta(x_t, t)$: 
\begin{equation}
\begin{split}
    x_{t} &= \sqrt{\bar{\alpha}_{t}} \left[ \bar{x}_{t-1} + \left( \gamma_t - \gamma_{t-1} \right) \epsilon_\theta(x_t, t) \right] \\
    &\approx \sqrt{\bar{\alpha}_{t}} \left[ \bar{x}_{t-1} + \left( \gamma_t - \gamma_{t-1} \right) \epsilon_\theta(x_{t-1}, t-1) \right].
\end{split}
\label{eq_approximation}
\end{equation}

This approximation enables the inversion of the DDIM process, allowing one to backtrack through the generation steps and recover the latent codes corresponding to a given output. Such inversion capabilities are valuable for editing tasks and other applications that require interpreting and manipulating intermediate representations.

\section{Methods}

\begin{figure*}
\centering
\includegraphics[width=1.0\textwidth]{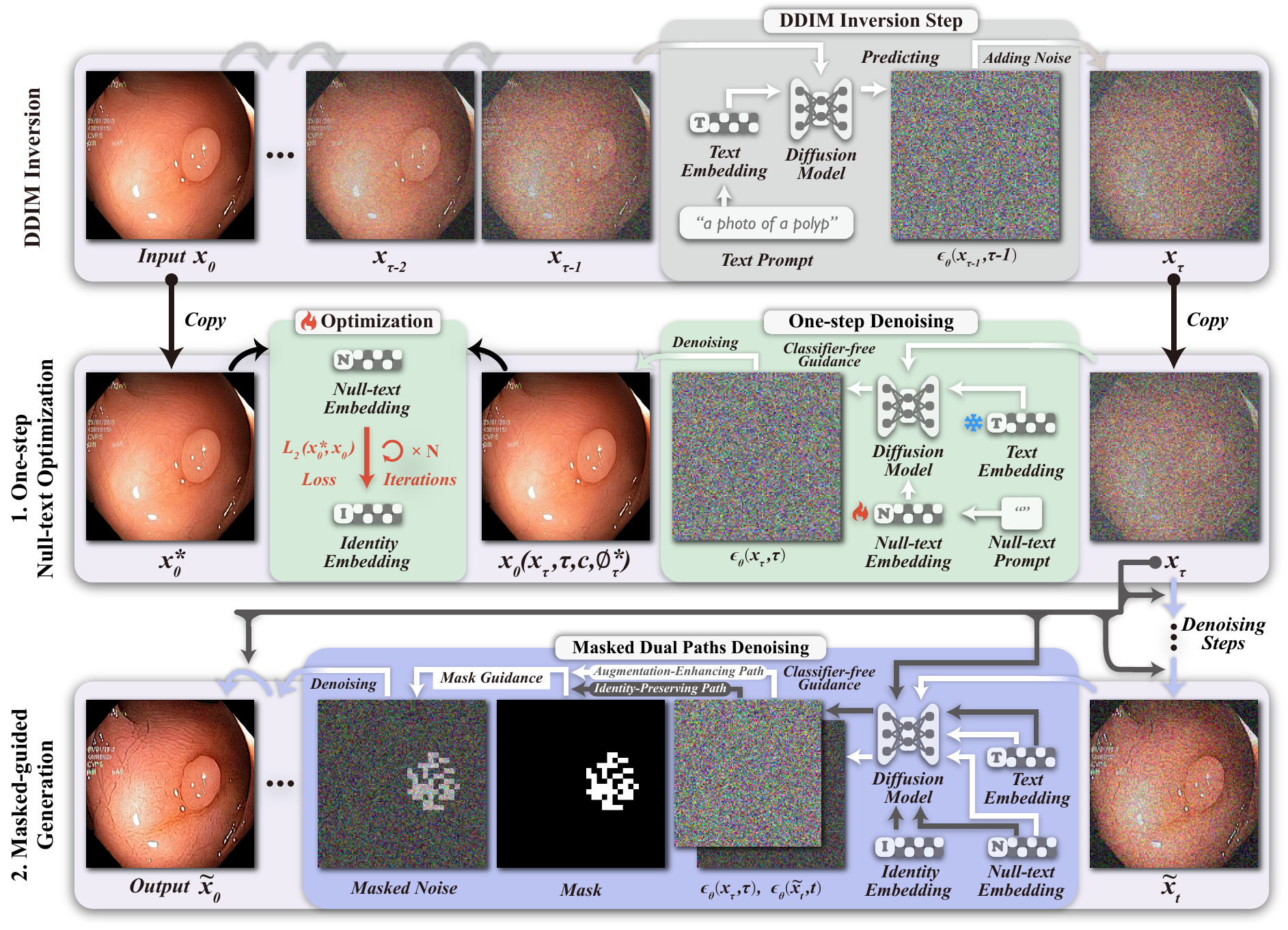}
\caption{The proposed pipeline of three key workflows are presented. The test image is processed through a sequence of steps to generate a noise image at a designated step count. Using this noise image, a one-step denoising process is employed to refine a trainable null-text embedding, enabling the stable generation of results that closely resemble the initial image. In the augmentation generation phase, this null-text embedding, guided by semantic and regional information, is leveraged to produce a series of augmented images.} 
\label{fig-2}
\end{figure*}

In this section, we detail our proposed TTGA framework. We introduce a multi-condition guidance approach to simultaneously enforce semantic coherence and preserve essential identity attributes. We further contribute a novel masked null-text inversion method tailored specifically for medical image editing, enabling precise, localized augmentation while maintaining clinical relevance and realism. The main workflow of this study is illustrated in Fig.~\ref{fig-2}

\subsection{One-step null-text optimization}
While image-to-image DDIM inversion combined with classifier-free guidance can generate images aligned with given conditions, the resulting edits often introduce significant random changes even in regions that should remain unchanged. Recent approaches tackle this issue by fine-tuning conditional \cite{dong2023prompt} or unconditional \cite{mokady2023null} embeddings. These approaches help preserve the more original content of the image during the prediction process guided with new information. However, in medical images, detailed textual conditions are rarely available, as typical datasets offer only limited categorical or semantic labels. Consequently, we require an editing method that does not depend heavily on precise textual prompts.

\subsubsection{Null-text inversion}
The vanilla null-text inversion \cite{mokady2023null} learns a new set of null-text embeddings during the DDIM inversion process, ensuring that the noise predicted at each step approximates the forward-added noise distribution. This procedure reconstructs samples that remain highly consistent with the original input. Specifically, pivotal tuning is executed sequentially from $x_T$ to $x_0$:
\begin{equation}
\mathcal{L}_t=\| x_{t-1}^* - x_{t-1}(x_t, t, c, \varnothing_t^*) \|_2^2,
\end{equation}
where $\bigl\{x_t^*\bigr\}_{t=0}^T$ is a series of sample templates obtained by DDIM inversion. Only the null-text embedding $\varnothing_t^*$ is updated in the optimization at the step $t$. Once the conditional embedding $c$ remains fixed, using the optimized null-text embeddings $\bigl\{\varnothing_t^*\bigr\}_{t=1}^T$ as unconditional embeddings ensures that the reconstructed sample closely resembles the original $x_0^*$.

Although the objective of null-text inversion is to reconstruct the test sample and subsequently perform editing based on new prompts, the scarcity of well-pretrained text-to-image models in the medical domain makes this approach difficult. As an improvement, we propose a \emph{mask-guided} strategy (see Sec.~\ref{sec_4.2}), which assigns varying augmentation strengths to different regions of the image. This strategy preserves the high realism of augmented images without relying on complex conditional embeddings.

\subsubsection{One-step optimization} \label{one_step}
The original null-text inversion requires step-wise optimization of the null-text embeddings across all timesteps in the DDIM scheme, in order to enable fine-grained semantic control starting from a fully noisy input $x_T$. However, such richly guided embeddings are redundant for generative augmentation, where the objective is not semantic editing but rather preserving the original image’s appearance. Moreover, this full-step optimization is computationally intensive. To address this, we modify the null-text optimization process to better suit our needs. Specifically, since our goal is only to retain the visual appearance of the input without performing text-level control during generation, we make two key changes: 1) we initiate the process from a single intermediate DDIM timestep $\tau$, and 2) we perform a one-step inversion from $\tau$ to $0$, instead of optimizing across all steps from $T$ to $0$:
\begin{equation}
\mathcal{L}_\tau=\| x_0^* - x_0(x_\tau, \tau, c, \varnothing_\tau^*) \|_2^2,
\label{one_step_optimization}
\end{equation}
where only the null-text embedding $\varnothing_\tau^*$ at timestep $\tau$  is updated during optimization. This process is illustrated in the center flow of Fig.~\ref{fig-2}. It significantly improves the efficiency of both the optimization and denoising processes while still preserving the content of the original image.

\subsection{Masked-guided generation} \label{sec_4.2}
\subsubsection{Multi-condition guidance}
Classic classifier-free guidance considers a single condition $c$ to guide the sample generation. However, many applications, such as generative augmentation in medical imaging, require enforcing multiple constraints concurrently. For instance, one might need to ensure that the synthesized image satisfies a semantic condition $c$ (e.g., specifying the presence or characteristics of particular lesions) while preserving identity-related attributes captured by $r$ (e.g., patient-specific anatomical details).

To incorporate both semantic and identity conditions, we introduce two guidance scales: $\lambda_c \geq 0$ for the semantic condition $c$ and $\lambda_r \geq 0$ for the identity condition $r$. Building on the single-condition setup in Eq.~\ref{sde-single}, we can introduce multiple conditions by applying the chain rule of conditional probabilities. Then we extend it to handle both conditions using separate guidance scales to balance the unconditional baseline, the content condition, and the identity constraint:
\begin{equation}
\begin{split}
    \nabla_{x_t} \log p(x_t \mid c, r) \approx \, &(1 - \lambda_c) \nabla_{x_t}\log p(x_t) \\
    &+ \lambda_c \nabla_{x_t}\log p(x_t \mid c) \\
    &+ \lambda_r \nabla_{x_t}\log p(r \mid x_t, c),
\end{split}
\label{eq_combined_guidance_3line}
\end{equation}
where the semantic term $\nabla_{x_t}\log p(x_t \mid c)$ guides the image toward desired semantic features, and the identity-related term $\nabla_{x_t}\log p(r \mid x_t, c)$ refines the sample to incorporate the identity-related attributes.

To realize this multi-condition guidance in practice, we extend classifier-free guidance to combine noise predictions under three conditions: unconditional ($\varnothing$), semantic-only ($c$), and semantic-plus-identity ($c, r$). By linearly mixing these predictions, we obtain:
\begin{equation}
\begin{split}
    \ddot{\epsilon}_\theta(x_t,t,\varnothing,r,c) = \, &\epsilon_\theta(x_t,t,\varnothing) \\
    &+ \lambda_c\bigl[\epsilon_\theta(x_t,t,c)-\epsilon_\theta(x_t,t,\varnothing)\bigr] \\
    &+ \lambda_r\bigl[\tilde{\epsilon}_\theta(x_t,t,r,c)-\epsilon_\theta(x_t,t,c)\bigr],
\end{split}
\label{classifier_free_multi}
\end{equation}
where $\epsilon_\theta(x_t,t,c)$ represents the noise prediction under the semantic content alone, and $\tilde{\epsilon}_\theta(x_t,t,r,c)$ denotes the prediction when both identity and semantic constraints are applied. The other terms maintain the same interpretation as in Eq.~\ref{classifier-free-single}. By adjusting $\lambda_c$ and $\lambda_r$, this multi-condition framework can simultaneously preserve key identity attributes and enforce semantic coherence, making it well-suited for generative augmentation tasks where multiple constraints must be satisfied in tandem. 

\subsubsection{Dual denoising paths with masks}\label{sec_dual_path}
DDIM can estimate noise across arbitrary step intervals. As shown in Eq.~\ref{ddim_sample_delta}, samples from larger time steps can be used to predict smaller time-step samples. Thus, multiple large time steps can be combined to form a new, denoised sample. Leveraging this property, we combine two DDIM samples using spatial masks that split the image into regions requiring different treatments. Given a time step $t$ and an interval $\Delta t$, we define two intermediate intervals $\Delta \spadesuit$ and $\Delta \clubsuit$ between $0$ and $\Delta t$. By merging the samples at these two steps, we can construct:
\begin{equation}
\begin{split}
    \bar{x}_{t - \Delta t} &= \, M_{t - \Delta t}^\spadesuit \bigl[ \bar{x}_{t - \Delta \spadesuit} + (\gamma_{t - \Delta t} - \gamma_{t - \Delta \spadesuit}) \epsilon_\theta(x_{t - \Delta \spadesuit}, t - \Delta \spadesuit) \bigr] \\
    &+ M_{t - \Delta t}^\clubsuit \left[ \bar{x}_{t - \Delta \clubsuit} + (\gamma_{t - \Delta t} - \gamma_{t - \Delta \clubsuit}) \epsilon_\theta(x_{t - \Delta \clubsuit}, t - \Delta \clubsuit) \right],
\end{split}
\label{dual_denoise}
\end{equation}
where $M_{t - \Delta t}^\spadesuit$ and $M_{t - \Delta t}^\clubsuit$ are binary masks that partition the image into regions treated differently during the denoising process. This framework facilitates realistic image generation with localized modifications, providing an effective solution for medical image editing.

To adapt this generic formulation for our task, we aim to generate realistic images with localized changes based on the original image without relying on modifications to the conditional embeddings. We specify the roles of the two paths. The \textit{identity-preserving path} $\spadesuit$ focuses on retaining original details, while the \textit{augmentation-enhancing path} $\clubsuit$ introduces controlled randomness to the generated samples. The flow diagram at the bottom of Fig.~\ref{fig-2} illustrates the mask-guided denoising strategy.

\subsubsection{Identity-preserving paths}
For the \textit{identity-preserving path}, we adopt the proposed one-step null-text optimization scheme to retain identity-specific information. Following Eq.~\ref{one_step_optimization}, we fix the denoising step $t - \Delta \spadesuit$ to a constant $\tau$. During the denoising process, regions requiring content preservation rely on the single-step null-text inversion for noise prediction. The result of step $t$ of the identity-preserving denoising process $\bar{x}_{t-1}^\spadesuit$ is:
\begin{equation}
\bar{x}_{t-1}^\spadesuit = \bar{x}_\tau + (\gamma_{t - 1} - \gamma_\tau) \dot{\epsilon}_\theta(x_\tau,\tau,\varnothing_\tau^*,c),
\label{dual_denoise_spade}
\end{equation}
where $\dot{\epsilon}_\theta(x_\tau,\tau,\varnothing_\tau^*,c)$ follows the same single-step noise prediction approach as in Eq.~\ref{one_step_optimization} for single-step denoising. In this work, we employ the single-condition classifier-free guidance denoising method described in Eq.~\ref{classifier-free-single} at the time step $\tau$, replacing the null-text term $\varnothing$ with the optimized $\varnothing_\tau^*$ to preserve identity information.

\subsubsection{Augmentation-enhancing paths}
For the \textit{augmentation-enhancing path}, we introduce controlled randomness to the generated samples while retaining essential identity information such as layout and style. This path imposes a dual-conditional constraint. The outcome of step $t$ in the augmentation-enhancing denoising process $\bar{x}_{t-1}^\clubsuit$ is:
\begin{equation}
\bar{x}_{t-1}^\clubsuit = \bar{x}_t + (\gamma_{t - 1} - \gamma_t) \ddot{\epsilon}_\theta(x_t,t,\varnothing,\varnothing_\tau^*,c),
\label{dual_denoise_club}
\end{equation}
where $\ddot{\epsilon}_\theta(x_t,t,\varnothing,\varnothing_\tau^*,c)$ incorporates both the semantic condition $c$ and the identity condition derived from the optimized $\varnothing_\tau^*$. Specifically, combining Eq. \ref{classifier_free_multi} with the dual constraints, we obtain a multi-conditional noise prediction:
\begin{equation}
\begin{split}
    \ddot{\epsilon}_\theta(x_t,t,\varnothing,\varnothing_\tau^*,c) = \, &\epsilon_\theta(x_t,t,\varnothing) \\
    &+ \lambda_c \bigl( \epsilon_\theta(x_t,t, c) - \epsilon_\theta(x_t,t,\varnothing) \bigr) \\
    &+ \lambda_r(1 - \omega) \bigl( \epsilon_\theta(x_t,t,\varnothing_\tau^*) - \epsilon_\theta(x_t,t,c) \bigr),
\end{split}
\label{multi_classifier_free_final}
\end{equation}
where we combine $\epsilon_\theta(x_t,t,\varnothing_\tau^*)$ and 
$\epsilon_\theta(x_t,t,c)$ to perform noise prediction for the dual-conditioned term $\tilde{\epsilon}_\theta(x_t,t,r,c)$. Specifically, we utilize the classifier-free guidance approach described in Eq.~\ref{classifier-free-single} at the time step $t$, while replacing the original $\varnothing$ with the $\varnothing_\tau^*$ to incorporate identity information. The noise prediction is determined by three noise components and their respective coefficients. Here, $\omega$ is fixed during null-text optimization, $\lambda_r$ controls the degree of identity preservation, and $\lambda_c$ and $\lambda_r$ together determine the strength of the semantic term.

\subsubsection{Mask derivation for medical image editing}\label{sec_mask}
By merging the two denoising paths, we form a dual-path denoising framework based on Eq.~\ref{dual_denoise} that allows localized control over image editing:
\begin{equation}
\bar{x}_t = M_t^\spadesuit \bar{x}_t^\spadesuit + M_t^\clubsuit \bar{x}_t^\clubsuit,
\label{dual_denoise_final}
\end{equation}
where $M_t^\spadesuit$ and $M_t^\clubsuit$ are spatial masks that determine which regions follow the \emph{identity-preserving path} and which follow the \emph{augmentation-enhancing path}. The choice of these masks governs the balance between preserving the original image content and introducing controlled variations. Below, we describe three simple yet effective mask generation schemes designed to accommodate diverse editing requirements.

\textbf{Bernoulli Scheme:}
In the simplest scenario, masks can be assigned randomly. We randomly initialize a mask $M_B \sim\mathcal{B}(s_m,p_m)$ that follows a Bernoulli distribution, where $s_m$ is the size of $x_t$ and $p_m$ represents the probability of taking the value $1$. This yields $M_t^\spadesuit = M_B$ and $M_t^\clubsuit = 1 - M_B$. Here, setting a higher $p_m$ favors preserving more of the original content (assigning more pixels to the identity-preserving path), while a lower $p_m$ allows for broader augmentation.

\textbf{Attention Scheme:}
While the Bernoulli scheme is agnostic to image content, medical image editing often requires more precise control. Following the prompt-to-prompt framework \cite{hertz2022prompt}, we can utilize attention maps derived from the diffusion model’s internal representation. These attention maps highlight condition-relevant or lesion-related regions, helping identify areas that should remain faithful to the original structure. Let $M_P(x_t)$ be a function that outputs a mask from the attention-based heatmap, assigning 1 to pixels in identity-critical regions and 0 elsewhere. We then set: $M_t^\spadesuit = M_P(x_t)$ and $M_t^\clubsuit = 1 - M_P(x_t)$. In this way, crucial anatomical or lesion areas are preserved by the identity-preserving path, avoiding distortions that could undermine clinical usefulness. Meanwhile, the non-critical background regions are assigned to the augmentation-enhancing path, allowing for increased variation and diversity.

\textbf{Hybrid Scheme:}
While the attention scheme can provide more meaningful spatial control, it may sometimes be desirable to introduce additional complexity or randomness. The hybrid scheme combines both the Bernoulli and attention-based approaches, interweaving identity-preserving and augmentation-enhancing regions. This is achieved by randomly mixing portions of the identity and augmentation masks: 
\begin{equation}
\left\{
\begin{array}{ll}
M_t^\spadesuit = M_B M_P(x_t) + (1-M_B) (1-M_P(x_t)) \\
M_t^\clubsuit = M_B (1-M_P(x_t)) + (1-M_B) M_P(x_t)
\end{array}
\right..
\label{mask_hybrid}
\end{equation}
In other words, the hybrid scheme selectively incorporates segments of the attention-based mask into the Bernoulli-generated regions. The resulting flexibility makes it suitable for scenarios where strict identity preservation in some areas is required, but a degree of random variability is also beneficial for augmentation and realism.

\begin{algorithm}[htb]
\caption{Masked Null-Text Inversion Generation}
\label{alg_1}
\begin{algorithmic}[1]
\REQUIRE ~~\\
An input sample $x_0$, and a semantic condition $c$; \\ 
An intermediate inversion step $\tau$; \\ 
Guidance scales $\omega$, $\lambda_c$, and $\lambda_r$; \\ 
Mask generation functions $\bigl\{ M_t^\spadesuit \bigr\}_{t=0}^{\tau-1}$ and $\bigl\{ M_t^\clubsuit \bigr\}_{t=0}^{\tau-1}$ that partition the samples into identity-preserving regions and augmentation-enhancing regions, respectively. 
\ENSURE ~~\\
An augmented image $\tilde{x}_0$ that preserves key identity attributes while introducing controlled semantic variations.

\STATE Perform DDIM inversion from $x_0$ up to step $\tau$, obtaining the intermediate latents $\left\{x_t^*\right\}_{t=1}^\tau$;

\STATE Initialize $x_\tau \gets x_\tau^*$ for null-text optimization and denoising loop;

\STATE Conduct single-step null-text optimization at step $\tau$ to yield the tuned null-text embedding $\tilde{\varnothing}_\tau$, ensuring $x_0(x_\tau, \tau, c, \varnothing_\tau^*) \approx x_0^*$;

\FOR{$t = \tau$ \textbf{down to} $1$}
    \STATE $\epsilon_t^\spadesuit \gets \dot{\epsilon}_\theta(x_\tau,\tau,\varnothing_\tau^*,c,\omega)$;
    
    \STATE $\tilde{x}_{t-1}^\spadesuit \gets Denoise(x_\tau, \epsilon_t^\spadesuit) $;

    \STATE $\epsilon_t^\clubsuit \gets \ddot{\epsilon}_\theta(\tilde{x}_t,t,\varnothing,\varnothing_\tau^*,c; \left\{ \omega, \lambda_c, \lambda_r \right\} )$;
    
    \STATE $\tilde{x}_{t-1}^\clubsuit \gets Denoise(\tilde{x}_t, \epsilon_t^\clubsuit) $;

    \STATE Determine $M_{t-1}^\spadesuit$ and $M_{t-1}^\clubsuit$ at time step $t-1$;
    
    \STATE $\tilde{x}_{t - 1} \gets Blend(\tilde{x}_{t-1}^\spadesuit, M_{t-1}^\spadesuit, \tilde{x}_{t-1}^\clubsuit, M_{t-1}^\clubsuit)$;
\ENDFOR
\RETURN $\tilde{x}_0$
\end{algorithmic}
\end{algorithm}

To formalize the integration of these mask schemes into the generative augmentation process, we present Alg.~\ref{alg_1} Masked Null-Text Inversion Generation which delineates a systematic procedure for generating augmented images. Specifically, it employs DDIM inversion to obtain intermediate latents, optimizes a single-step null-text embedding to align with the original image, and applies dual denoising paths guided by the spatial masks derived from the aforementioned schemes. 

\subsection{Generative conditions in medical context}

The generative framework, as detailed in Sec. ~\ref{sec_4.2}, utilizes a multi-condition guidance approach involving an abstract "semantic condition $c$" and "identity condition $r$". A practical challenge in applying this framework to medical imaging is that, unlike general-purpose image datasets, medical data often lacks rich, descriptive text labels. This section clarifies more details of adapting the generative backbone to the medical domain and provides concrete definitions for these two conditions as implemented in our experiments. We employ Low-Rank Adaptation (LoRA) \cite{hu2022lora} to adapt the existing generative backbone which was pre-trained on general-domain images (e.g., LAION-5B). We fine-tune the model on the training set corresponding to each specific downstream segmentation task. This process aligns the generative model with the specific visual distribution of the medical domain (e.g., fundus or polyp images) and associates it with simple, class-level text prompts. This domain adaptation is crucial for ensuring that all generated augmentations are "in-distribution" and medically plausible, and it preserves the classifier-free guidance mechanism without requiring complex prompt engineering.

\subsubsection{The semantic condition as a domain anchor}
We explicitly define the semantic condition $c$ as the CLIP embedding of a simple, class-level text prompt. For example, $c = \text{CLIP}(\text{"a photo of a polyp"})$ for the polyp datasets.We clarify that the primary goal of TTGA is not to perform complex, text-driven editing. Instead, the function of $c$ is to act as a "Domain Anchor" during the classifier-free guidance process. It steers the generation towards the specific medical image distribution learned during the LoRA fine-tuning. For our objective generating plausible perturbations rather than novel semantic content (a simple class) level prompt is sufficient to anchor the generation within the correct medical domain.

\subsubsection{The identity condition as an image-specific container}
We define the identity condition $r$ not as a text prompt, but as the optimized null-text embedding $\varnothing_\tau^*$ obtained from our proposed one-step null-text optimization (detailed in Sec.~\ref{one_step}).This definition is based on the principle of classifier-free guidance, which operates by interpolating between a conditional prediction (guided by $c$) and an unconditional prediction (guided by a null-text embedding $\varnothing$). The original null-text inversion method demonstrated that by optimizing this $\varnothing$ embedding, the generative process can be guided to precisely reconstruct a specific source data $x_0$. We adapt this concept: at test time, for each input $x_0$, we perform the one-step optimization to find a unique $\varnothing_{\tau}^{*}$ that encodes the specific identity and content of $x_0$. This image-specific $\varnothing_{\tau}^{*}$ becomes the identity condition $r$, effectively capturing the patient-specific anatomical details of the input sample.

\subsubsection{Re-contextualizing the dual denoising paths}
With $c$ and $r$ now clearly defined, we can provide evidence for the function of the dual denoising paths (from Sec.~\ref{sec_dual_path}).

\textbf{Identity-Preserving Path}: This path is guided by $r = \varnothing_{\tau}^{*}$. By its very definition, $\varnothing_{\tau}^{*}$ is the embedding optimized to perfectly reconstruct the original data $x_0$. Therefore, this path inherently preserves the fine-grained, original details of the image.

\textbf{Augmentation-Enhancing Path}: This path combines $c$ and $r$. The identity condition $r$ (as $\varnothing_{\tau}^{*}$) ensures the generated output retains the core identity and structure of $x_0$. Simultaneously, the semantic condition $c$ introduces domain-specific, plausible randomness (e.g., variations in texture or lighting) that was learned during the LoRA domain adaptation.

\subsection{Test-time generative augmentation}
We further introduce test-time generative augmentation (TTGA), building on our generative scheme. Given a test image $x^{'}$, a set of null-text embeddings $E^{'}$ is obtained through one round of null-text optimization. Subsequently, $N$ masks are generated and applied through $N$ rounds of masked null-text inversion, resulting in a set of generative augmented images. A segmentation model performs multiple segmentations on the augmented images, and an ensemble of these segmentation results is used to obtain a posterior regarding the test image:
\begin{equation}
p(x^{'}) = \frac{1}{N} \cdot \sum_{i=1}^N p(f(x^{'},m^{'}_i,E^{'})),
\end{equation}
where the $f(\cdot)$ is the execution of a masked null-text inversion. With regard to the ensemble results, their uncertainty $H$ can be further estimated. In this work, we employ entropy as a measure of uncertainty:
\begin{equation}
H(p)=\sum_{k=1}^{K} p_k \cdot log_2(p_k),
\label{entropy}
\end{equation}
where the $K$ is the quantity of different classes.

\section{Experiments}
\subsection{Materials}\label{Materials_sec}
\textbf{1) Datasets:} We evaluate our TTGA in three medical image segmentation tasks. This evaluation demonstrates the method's general performance. We use two main validation strategies. The first strategy uses nnU-Net-v2\footnote{Official Code of nnU-Net-v2 for Medical Images Segmentation. \url{https://github.com/MIC-DKFZ/nnUNet/}} which is a widely accepted baseline in medical imaging. This model provides a consistent architecture for our evaluations. The second strategy uses existing, specialized segmentation models. These models have publicly available weights and are designed for specific tasks:

\begin{itemize}
\item \textbf{Optic Disc and Cup Segmentation}\footnote{Official Code of Segmentation Model used in Optic Disc and Cup Segmentation. \url{https://github.com/askerlee/segtran/}}: We first employ the REFUGE dataset to evaluate the proposed method on segmentation of optic discs and cups in images acquired by fundus cameras. We employ a SOTA Transformer-based segmentation model SegTran~\cite{li2021medical}, with utilization of the well-trained model weights. The dataset has $400$ training images, $400$ validation images, and $400$ test images. We follow SegTran and resize all images to $288 \times 288$ pixels.

\item \textbf{Polyp Segmentation}\footnote{Official Code of Segmentation Model used in Polyp Segmentation. \url{https://github.com/baiboat/HSNet/}}: We extend our evaluation to include polyp segmentation in endoscopic images. We employ HSNet~\cite{zhang2022hsnet} as our segmentation model, utilizing the codes and weights. The model was trained on $1450$ images. These images came from the Kvasir and CVC-ClinicDB datasets. All training images are resized to $352 \times 352$ pixels. The test set includes images from five datasets. The ``unseen'' part includes images from three other datasets. These are Kvasir ($100$ images), CVC-ClinicDB ($60$ images), CVC-ColonDB ($380$ images), CVC-300 ($60$ images), and ETIS-LiaribPolypDB ($196$ images).

\item \textbf{Skin Lesion Segmentation}\footnote{Official Code of Segmentation Model used in Skin Lesion Segmentation. \url{https://github.com/wurenkai/H-vmunet}}: 
We investigate the effectiveness of the proposed method in skin lesion segmentation. We use a specialized model from H-vmunet~\cite{wu2025h} for skin lesion segmentation. We use the provided pre-trained model. This model was trained on $1250$ training images and $150$ validation images. These images were selected from the ISIC 2017 training set. These images were resized to $256 \times 256$ pixels. We test the performance on three datasets. These are the ISIC 2017 test set ($600$ images), the ISIC 2018 test set ($1000$ images), and the PH2 dataset ($200$ images).
\end{itemize}

To ensure our experiments to be fair and avoid extra factors, we directly use the official open-source weights for the three specialized models. We train the nnU-Net-v2~\cite{isensee2021nnu} model ourselves. For this training, we strictly follow the same dataset splits and image sizes as the specialized models. We use all other nnU-Net-v2 parameters at their default settings. To ensure a fair comparison with state-of-the-art methods, we strictly followed the official data partitioning provided by the specialized models. Additionally, to evaluate the robustness of our method under varying data conditions (e.g., data scarcity and domain shift), we conducted extensive ablation studies on different data splits, which are detailed in Sec. S5 of the Supplementary Material.

\textbf{2) Metric:} In this study, we evaluate the effectiveness of the proposed testing strategy from two critical perspectives:
\begin{itemize}
\item \textbf{Medical Images Segmentation}: This aspect investigates whether the testing strategy enhances the overall quality of the segmentation results. An effective strategy should mitigate the impact of unstable perturbations (random or unpredictable variations that may compromise performance) thereby improving the generalization and reliability of the segmentation outcomes.

\item \textbf{Pixel-wise Error Estimation}: This evaluation assesses the testing strategy’s ability to predict uncertainty in segmentation outputs at the pixel level. Specifically, we measure the discrepancy between the segmentation network’s predictions on the original image and the corresponding ground truth. The primary objective is to determine whether the strategy can accurately localize and encompass regions of uncertainty within the segmentation.
\end{itemize}

During testing, we generated $N=10$ augmented samples for each test image. This quantity was empirically selected to achieve an optimal balance between segmentation accuracy gains and computational costs. A detailed analysis regarding the choice of augmentation quantity is provided in Sec. S2 of the Supplementary Material. The two aforementioned evaluations offer comprehensive insights into the effectiveness of the testing strategy from the perspectives of uncertainty prediction and segmentation quality. We systematically analyze performance using metrics tailored for each evaluation.

For Segmentation Accuracy, we employ the \textbf{Dice Similarity Coefficient (DSC)}, \textbf{Area Under the Curve (AUC)}, and the \textbf{95th percentile Hausdorff Distance (HD95)}. The DSC quantifies the spatial overlap between the prediction and the ground truth, where a higher value (ranging 0-100) indicates better overlap. The AUC (Area Under the Receiver Operating Characteristic Curve) assesses threshold-independent classification performance, where a higher value (ranging 0-100) signifies better discrimination. The HD95 is a boundary-based metric robust to outliers that measures the 95th percentile of the distance between the predicted and ground truth boundaries, where a lower value indicates closer boundary agreement. Together, these metrics evaluate spatial overlap (DSC), classification performance (AUC), and boundary fidelity (HD95), providing a comprehensive assessment of the segmentation quality.

For Pixel-wise Error Estimation, we utilize \textbf{DSC}, \textbf{AUC}, and the \textbf{Normalized Surface Distance (NSD)} (with a 1\% tolerance). We specifically selected NSD over HD95 for this task. The NSD is a surface-based metric that quantifies the percentage of the surface within an accepted tolerance (1\%), making it highly suitable for boundary-like structures; a higher value (ranging 0-100) is better. We selected NSD because pixel-wise errors typically manifest as thin, elongated regions along the segmentation target's boundaries. The morphology of these error regions is ill-suited for HD95, which assesses the distance between boundary surfaces. NSD is therefore used as a more appropriate measure to quantify the surface agreement for these specific error structures.

\subsection{Implementation details}

In this study, we employ the pre-trained Stable Diffusion v1-5 \cite{rombach2022high} as the backbone for generative diffusion model. For in-distribution adaptation, we employed the official LoRA template from Hugging Face, fine-tuning the diffusion model with the same dataset used for training the corresponding segmentation model. This process facilitates a rapid and lightweight adaptation of the generative model to the specific domain of medical segmentation images. To ensure higher image quality and maintain a uniform experimental configuration across all three tasks, we generated images at a $512 \times 512$ resolution during the generative phase. Subsequently, at inference time, these generated images were downsampled to the specific input dimensions required by each segmentation model. To achieve a balance between the randomness of the augmented content and the preservation of the overall layout, the denoising process begins at a timestep $\tau=300$, with a total of $T=1000$. During the DDIM Inversion phase, noise is incrementally added at intervals of $10$ steps until the starting timestep $\tau$ is reached. For the null-text optimization phase, we adopt a one-step sampling approach (from $\tau$ to 0) to refine the latent representation. This optimization minimizes the mean squared error (MSE) loss between the predicted $x_0^*$ and the latent representation of the original image $x_0$. We employ the Adam optimizer for this task, configured with a learning rate of $0.1$, a maximum of $500$ optimization steps, and an early stopping threshold of $5 \times 10^{-4}$. All experiments were conducted on a standard GPU workstation. Detailed hardware specifications, software environments, and a comprehensive runtime analysis are provided in Sec. S1 of the Supplementary Material.

\subsection{Parameter settings}\label{parameter_study}

\begin{figure*}
\centering
\includegraphics[width=1.0\textwidth]{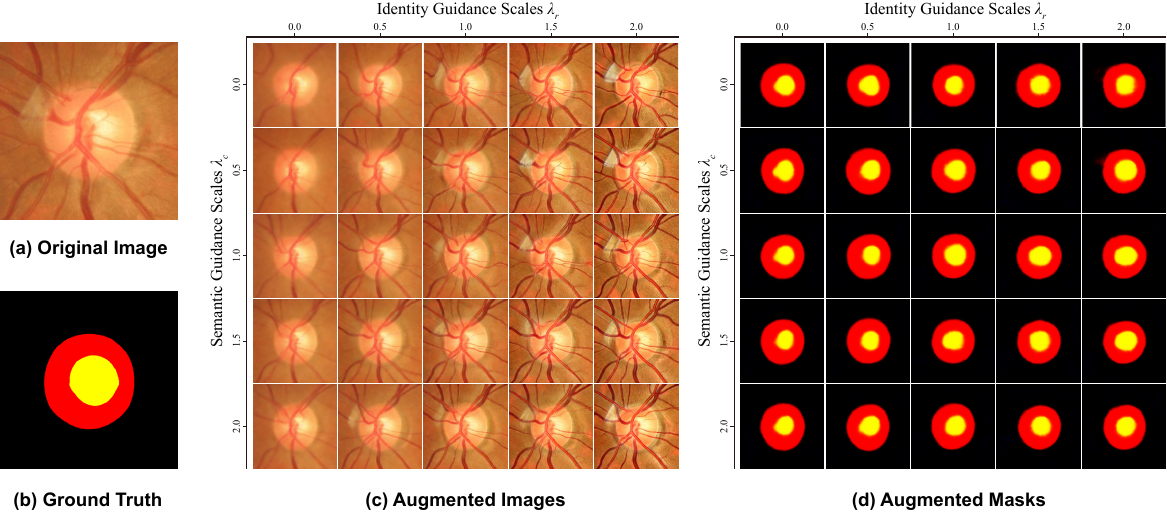}
\caption{Results of TTGA on a fundus image under different guidance scales. (a) The original, unaugmented image. (b) The corresponding ground truth segmentation of the optic disc and cup. (c) Visualization of augmented images generated using various combinations of identity guidance scales and semantic guidance scales. (d) Segmentation results produced by the model on the augmented images.} 
\label{fig-3}
\end{figure*}

In this subsection, we explore the influence of key parameters on the efficacy of our proposed TTGA method. Specifically, our approach leverages the augmentation-enhancing paths, a technique derived from the proposed masked null-text inversion framework, to execute generative augmentation on regions of test images. This process is modulated by two pivotal guidance parameters, as outlined in Eq.~\ref{multi_classifier_free_final}: 
\begin{itemize}
\item \textbf{Semantic Guidance Scale $\lambda_c$}: This parameter dictates the intensity of semantic alignment between the generated content and the provided text prompt, thereby influencing the relevance of the augmented output to the textual description.

\item \textbf{Identity Guidance Scale $\lambda_r$}: This parameter regulates the degree to which the augmented image preserves the identity-specific attributes of the original test image, ensuring similarity to its inherent characteristics. 
\end{itemize}

Through a detailed examination of these guidance parameters, we elucidate their respective contributions to achieving an optimal balance between semantic coherence and identity retention in the generated outputs. As depicted in Fig.~\ref{fig-3}, we applied our proposed TTGA method to a test fundus image (shown in Fig.~\ref{fig-3}(a)), producing augmented images under varying values of $\lambda_c$ and $\lambda_r$, with the results presented in Fig.~\ref{fig-3}(c). Complementing this qualitative analysis, we provide a detailed quantitative evaluation of these parameters in Sec. S2 of the Supplementary Material.

To evaluate their effects comprehensively, we systematically adjusted both parameters across a range from $0.0$ to $2.0$ and observed their influence on the content of the generated images. The qualitative findings indicate that when the semantic guidance scale $\lambda_c$ falls below $1.0$, the generated images frequently exhibit deviations from realistic structural distributions, manifesting as fragmented vessels, excessive curvature, or abnormal spatial arrangements. In contrast, when $\lambda_c$ is set to $1.0$ or higher, the generated images consistently display more pronounced characteristics typical of fundus images. Moreover, as outlined in Eq.~\ref{multi_classifier_free_final}, assigning $\lambda_c$ a value of $1.0$ nullifies the coefficient of the raw null-text noise term during the denoising process, simplifying computations by limiting the denoising operation to semantic and identity noise components alone. To streamline subsequent discussions and enhance computational efficiency, \textbf{we fixed $\lambda_c$ at $1.0$ for all further experiments in this section}.

Turning to the identity guidance scale $\lambda_r$, its influence on the generated image content is markedly more substantial than that of $\lambda_c$. Specifically, identity-related features (such as vessel coloration and optic disc edge definition) become progressively more prominent as $\lambda_r$ increases. When $\lambda_r$ is below $1.0$, certain identity details may be suppressed, resulting in the loss of some vessels or the introduction of spurious ones not present in the original image. Conversely, when $\lambda_r$ exceeds $1.0$, these identity features are amplified, occasionally leading to over-enhancement, such as intensified colors or overly sharpened edges. Consequently, varying $\lambda_r$ introduces a range of randomness in the intrinsic properties of the generated images, thereby yielding diverse augmentation outcomes. To achieve greater diversity in the generated augmented images while preserving the characteristics of the original image, \textbf{we configured the parameter $\lambda_r$ for each augmented image to be uniformly distributed around $1.0$ (with a radius of 0.5).} This choice is supported by the quantitative results in Fig. S1 of the Supplementary Material, which show that a radius of 0.5 yields the most robust boundary fidelity (HD95) while avoiding excessive variance.

The segmentation masks for the optic disc and cup, generated by our network under different parameter configurations, are illustrated in Fig.~\ref{fig-3}(d), revealing subtle variations in the segmentation results. On one hand, the generative augmentation impacts the target segmentation regions to differing degrees: internal textures and edge profiles exhibit parameter-dependent changes. Notably, since the generation process integrates information from the original image, the resulting augmentations remain controllable, centered around the original image as a baseline. This ensures that segmentation outcomes do not deviate unreasonably from the original results. On the other hand, for irrelevant background regions, TTGA employs semantically consistent generative augmentation. This approach maintains the overall realism and semantic alignment of the image content while introducing localized perturbations—such as alterations in vessel morphology overlapping the optic disc and cup, or modifications to background areas. These controlled disturbances enhance data variability, particularly in regions posing greater segmentation challenges, thereby providing additional uncertainty support to improve model robustness. Regarding the mask schemes, the hybrid scheme ($p_m=0.75$) was selected as the default configuration, as it consistently outperformed the Bernoulli and attention schemes in our extensive quantitative comparisons (see Fig. S1 in Supplementary Material).

\subsection{Qualitative comparisons}\label{Qualitative_sec}
\begin{figure*}
\centering
\includegraphics[width=1.0\textwidth]{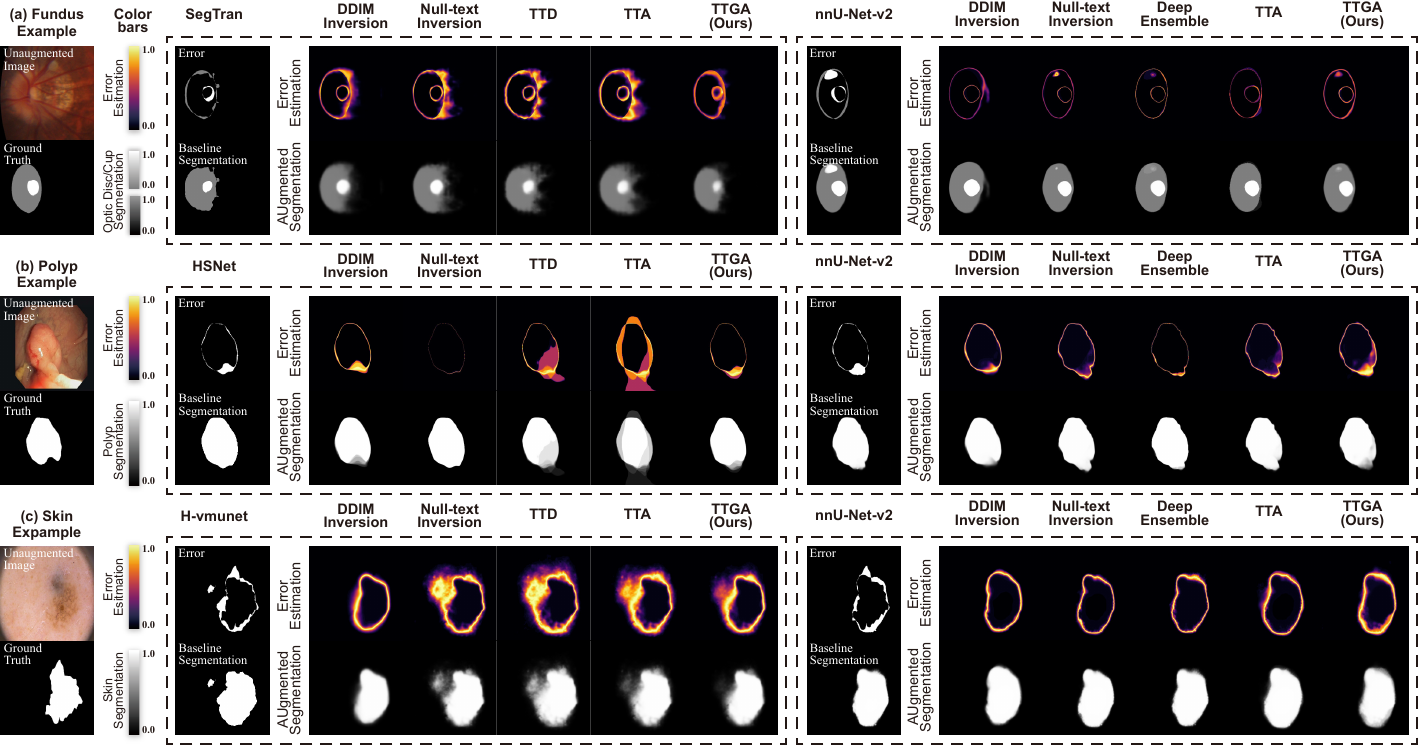}
\caption{Qualitative comparison of segmentation results and Error estimation. This figure provides a visual comparison of TTGA (Ours) against baseline models and other test-time methods across three tasks: (a) Fundus, (b) Polyp, and (c) Skin. The "Error" (target highlighted in orange) represents the ground truth for uncertainty which is generated by visualizing the pixel-wise difference between the binarized baseline segmentation (using a 0.5 threshold on the 0-1 normalized output) and the "Ground Truth" (also highlighted in orange). The objective for the "Segmentation" column is to match the "Ground Truth," while the objective for the "Error Estimation" column is to match the "Error." The color bars indicate the scales for segmentation probability and error estimation, respectively}
\label{qualitative1}
\end{figure*}

To intuitively evaluate the proposed framework, this section presents a qualitative comparison of our TTGA method against state-of-the-art test-time strategies, focusing on both the generated augmented images and the final segmentation outcomes.

We first re-introduce the comparative methods. The first strategy, \textbf{Test-time Dropout (TTD)}, generates perturbed model outputs by randomly deactivating specific layers during inference. Monte Carlo (MC) dropout \cite{gal2016dropout} is adopted as the implementation mechanism for TTD. To ensure a fair comparison, the dropout rate configuration yielding the most favorable results for TTD is selected. We also include the second strategy \textbf{Deep Ensemble}, implemented as Snapshot Ensembles \cite{huang2017snapshot}, a classic and widely-used technique in medical imaging. It is important to note the implementation constraints: TTD cannot be applied to the nnU-Net-v2 model due to its high degree of encapsulation. Furthermore, as our objective is to utilize the official open-source parameters of the three specialized segmentation models without retraining, deep ensemble was not applied to them. The third strategy, Test-time \textbf{Augmentation (TTA)}, perturbs data outputs by applying transformations to the input image. A combination of multi-view \cite{moshkov2020test} and morphological transformations \cite{ayhan2022test} is utilized as the comparative approach for TTA. Consistent with the proposed TTGA method, all strategies aggregate 10 distinct segmentation results derived from a single test image. As comparative/ablation results, we also provide results from \textbf{DDIM inversion}~\cite{song2020denoising}  (which lacks identity preservation control) and \textbf{null-text inversion}~\cite{mokady2023null} (which lacks the ability to generate diverse outputs) as the fourth and fifth. The soft segmentation output from the \textbf{unaugmented}  test image serves as the baseline. To visually assess the generative quality underlying these strategies, we provide a qualitative comparison of the augmented samples produced by TTGA, DDIM inversion, and null-text inversion in Fig. S2 of the Supplementary Material.

Fig. ~\ref{qualitative1} provides qualitative examples from the three segmentation tasks. For each task, an input image and its corresponding ground truth (GT) are shown. We present segmentation results from two distinct baselines: the original specialized network and the nnU-Net-v2. The pixel-wise discrepancy between a baseline's segmentation and the GT is defined as the "Error", which serves as the target for uncertainty estimation. Therefore, the comparative methods are evaluated on two complementary objectives: 1) the fidelity of the "Error Estimation" map to the actual "Error" map, and 2) the proximity of the aggregated "Segmentation" result to the "Ground Truth".

In the first example (fundus disc/cup segmentation), the disc's right boundary is ambiguous and obscured by foreground structures. The SegTran baseline produces an unsmooth boundary, indicating low confidence in this region. While other methods identify this uncertainty, they tend to overestimate it, leading to significant over-segmentation in their aggregated results. In contrast, TTGA, by introducing semantic guidance to the augmentation, effectively reduces this invalid uncertainty. For the nnU-Net-v2 baseline, a clear over-segmentation of the cup is visible. Notably, only null-text inversion and TTGA accurately highlight this specific error. However, the limited stochasticity of null-text inversion restricts its ability to identify uncertainty along other parts of the boundary.

In the second example (polyp segmentation), the boundary between the polyp and the surrounding tissue is indistinct. The HSNet baseline appears overfitted to the training data, causing methods like TTA (which relies on geometric transforms) to fail significantly. The model is also sensitive to parameter perturbations, enabling TTD to locate the correct error region but at the cost of introducing many false-positive uncertainties. DDIM inversion, lacking identity preservation, identifies the general error location but with a shape that poorly matches the true "Error" boundary. TTGA, by balancing identity and semantics, provides the most accurate delineation of this error. A similar error pattern is observed in the nnU-Net-v2 result. Despite the general insensitivity of nnU-Net-v2 to data perturbation, TTGA is the only method to register high uncertainty in this specific error region, demonstrating its superior augmentation quality.

In the third example (skin lesion segmentation), the lesion boundary is exceptionally blurry, posing a significant challenge. The H-vmunet baseline produces several large error regions. The excessive randomness of DDIM inversion results in a loss of contour details. Other methods tend to over-focus on these challenging areas, broadly overestimating the scope of the uncertainty. While the nnU-Net-v2 baseline is less sensitive to data perturbations than H-vmunet, TTGA consistently demonstrates superior coverage of the true uncertainty regions compared to all other methods.

\subsection{On improving pixel-wise error estimation}\label{pixel_wise_error_Estimation}

\begin{table*}[htbp]
\centering
\caption{Performance comparison of different methods on pixel-wise error estimation across three tasks, evaluated by the DSC, AUC and NSD between uncertainty estimation regions and segmentation error regions. The results are binarized at a threshold of 0.5 (for the 0-1 range) before contributing to the metric calculation. In each set of experimental results, the best-performing value is indicated in bold font, while the second-best is indicated underlined font.}\label{tab_error}
\setlength{\tabcolsep}{3pt}
\begin{tabular}{lcccccccccc}
\toprule
\multirow{2}{*}{Method}  & \multicolumn{3}{c}{Fundus} & \multicolumn{3}{c}{Polyp} & \multicolumn{3}{c}{Skin} \\
\cmidrule(lr){2-4} \cmidrule(lr){5-7} \cmidrule(lr){8-10} 
& DSC $\uparrow$ & AUC $\uparrow$ & NSD $\uparrow$ & DSC $\uparrow$ & AUC $\uparrow$ & NSD $\uparrow$ & DSC $\uparrow$ & AUC $\uparrow$ & NSD $\uparrow$ \\

\midrule
Baseline &  \multicolumn{9}{c}{nnU-Net-v2} \\
\cmidrule(lr){2-10}

Unaugmented & 28.5\scriptsize{$\pm$4.3} & 84.1\scriptsize{$\pm$8.0} & 84.1\scriptsize{$\pm$4.6} & \underline{41.7{\scriptsize$\pm$9.9}} & 94.5{\scriptsize$\pm$10.4} & \underline{92.1{\scriptsize$\pm$13.5}} & 36.2\scriptsize{$\pm$10.0} & 88.3\scriptsize{$\pm$12.8} & 82.1\scriptsize{$\pm$12.8} \\

DDIM Inv. & \underline{32.4\scriptsize{$\pm$4.9}} & \underline{87.4\scriptsize{$\pm$7.8}} & \underline{84.3\scriptsize{$\pm$5.4}} & 39.6\scriptsize{$\pm$13.4} & 94.7\scriptsize{$\pm$10.6} & 76.8\scriptsize{$\pm$23.4} & \underline{40.8\scriptsize{$\pm$9.9}} & \underline{90.6\scriptsize{$\pm$11.3}} & 71.8\scriptsize{$\pm$20.6} \\

Null-text Inv. & 28.5\scriptsize{$\pm$4.3} & 84.2\scriptsize{$\pm$8.0} & \underline{84.3\scriptsize{$\pm$3.5}} & 41.3\scriptsize{$\pm$10.4} & 94.2\scriptsize{$\pm$9.8} & 91.6\scriptsize{$\pm$15.0} & 36.3\scriptsize{$\pm$10.3} & 88.8\scriptsize{$\pm$12.7} & 81.4\scriptsize{$\pm$12.6} \\

Deep Ensemble & 27.9\scriptsize{$\pm$8.2} & 66.7\scriptsize{$\pm$5.3} & 83.9\scriptsize{$\pm$5.8} & 37.4\scriptsize{$\pm$7.8} & 78.6\scriptsize{$\pm$9.1} & 91.0\scriptsize{$\pm$14.7} & 27.8\scriptsize{$\pm$10.7} & 66.8\scriptsize{$\pm$9.7} & \underline{84.3\scriptsize{$\pm$13.5}} \\

TTA & 28.8\scriptsize{$\pm$7.7} & 72.2\scriptsize{$\pm$8.0} & 83.9\scriptsize{$\pm$6.1} & 39.9\scriptsize{$\pm$9.7} & \underline{97.2\scriptsize{$\pm$5.1}} & 90.9\scriptsize{$\pm$14.6} & 39.3\scriptsize{$\pm$10.2} & 90.3\scriptsize{$\pm$11.8} & 79.4\scriptsize{$\pm$11.8} \\

\rowcolor[gray]{0.9} TTGA (\textbf{Ours}) & \textbf{34.5\scriptsize{$\pm$4.5}} & \textbf{90.3\scriptsize{$\pm$6.6}} & \textbf{84.5\scriptsize{$\pm$5.4}} & \textbf{43.5\scriptsize{$\pm$7.7}} & \textbf{97.5\scriptsize{$\pm$4.9}} & \textbf{93.4\scriptsize{$\pm$12.0}} & \textbf{43.8\scriptsize{$\pm$9.1}} & \textbf{92.4\scriptsize{$\pm$10.9}} & \textbf{85.8\scriptsize{$\pm$12.0}} \\

\midrule
Baseline & \multicolumn{3}{c}{SegTran} & \multicolumn{3}{c}{HSNet} & \multicolumn{3}{c}{H-vmunet} \\
\cmidrule(lr){2-4} \cmidrule(lr){5-7} \cmidrule(lr){8-10}

Unaugmented & 46.2\scriptsize{$\pm$6.5} & 96.9\scriptsize{$\pm$6.5} & 84.6\scriptsize{$\pm$11.5} & 10.3\scriptsize{$\pm$7.5} & 80.5\scriptsize{$\pm$8.6} & \underline{90.3\scriptsize{$\pm$11.2}} & 43.2\scriptsize{$\pm$8.6} & 90.6\scriptsize{$\pm$10.8} & 63.2\scriptsize{$\pm$29.0} \\

DDIM Inv. & 46.2\scriptsize{$\pm$16.3} & 97.0\scriptsize{$\pm$2.7} & 82.1\scriptsize{$\pm$10.6} & 22.2\scriptsize{$\pm$17.1} & 78.3\scriptsize{$\pm$15.3} & 74.4\scriptsize{$\pm$26.6} & 41.8\scriptsize{$\pm$9.4} & 89.9\scriptsize{$\pm$11.3} & 59.8\scriptsize{$\pm$29.6} \\

Null-text Inv. & 48.9\scriptsize{$\pm$14.6} & 97.6\scriptsize{$\pm$2.1} & 85.1\scriptsize{$\pm$9.8} & 15.7\scriptsize{$\pm$11.1} & 73.1\scriptsize{$\pm$7.5} & 89.6\scriptsize{$\pm$11.3} & 42.0\scriptsize{$\pm$9.2} & 90.3\scriptsize{$\pm$9.5} & 61.5\scriptsize{$\pm$32.4} \\

TTD & \underline{52.0\scriptsize{$\pm$12.3}} & 97.9\scriptsize{$\pm$2.1} & \underline{85.7\scriptsize{$\pm$10.5}} & \textbf{39.6\scriptsize{$\pm$16.5}} & 88.5\scriptsize{$\pm$11.0} & 85.1\scriptsize{$\pm$19.9} & 36.2\scriptsize{$\pm$9.5} & 84.2\scriptsize{$\pm$10.2} & 57.3\scriptsize{$\pm$31.0} \\

TTA & 51.3\scriptsize{$\pm$13.8} & \underline{98.1\scriptsize{$\pm$2.1}} & 82.1\scriptsize{$\pm$12.5} & 25.9\scriptsize{$\pm$15.8} & \underline{89.5\scriptsize{$\pm$10.2}} & 80.8\scriptsize{$\pm$18.3} & \underline{43.8\scriptsize{$\pm$8.9}} & \underline{91.6\scriptsize{$\pm$10.5}} & \underline{63.5\scriptsize{$\pm$30.3}} \\

\rowcolor[gray]{0.9} TTGA (\textbf{Ours}) & \textbf{53.6\scriptsize{$\pm$14.0}} & \textbf{98.6\scriptsize{$\pm$0.9}} & \textbf{86.5\scriptsize{$\pm$10.4}} & \underline{39.3\scriptsize{$\pm$14.6}} & \textbf{90.9\scriptsize{$\pm$9.6}} & \textbf{92.0\scriptsize{$\pm$12.0}} & \textbf{44.3\scriptsize{$\pm$8.4}} & \textbf{92.7\scriptsize{$\pm$9.9}} & \textbf{63.9\scriptsize{$\pm$28.9}} \\

\bottomrule
\end{tabular}
\end{table*}

The uncertainty estimation performance of comparisons strategies mentioned in Sec.~\ref{Qualitative_sec} is evaluated, with the results for pixel-wise error estimation presented in Tab.~\ref{tab_error}. We evaluate the performance on two distinct baselines for each task: a nnU-Net-v2 as mentioned in Sec.~\ref{Materials_sec}, and a task-specific segmentation model with open-source parameters. Due to the differences in parameter scales, comparisons between baselines are not drawn; we only compare our method and the comparative methods within the same baseline. The baseline of unaugmented uncertainty estimation is obtained from the soft segmentation output of the model applied to the original, unperturbed image. Overall, the proposed TTGA achieves the best performance in all metrics on the nnU-Net-v2, and either the best or second-best performance on the specialized models across all tasks, demonstrating its consistent capability to deliver satisfactory outcomes. The standard deviations of the metrics are also reported, where TTGA exhibits consistently low values, indicating the absence of high performance fluctuations and demonstrating robustness in pixel-wise error estimation. Owing to the limited controllability of DDIM inversion and the reduced flexibility of null-text inversion, their pixel-wise error estimations tend to converge toward or fall below that of the baseline, reflecting constrained expressiveness. In contrast, the comparative methods (TTD and TTA) exhibit task-specific adaptability challenges, occasionally underperforming significantly relative to other approaches. Notably, even when TTGA does not secure the top position, its performance remains closely competitive with the best-performing method.

Overall, TTGA achieves the best performance across most tasks. Although its DSC on the polyp pixel-wise error estimation task is slightly lower than that of TTD, a scenario we further investigate in the mixture solution analysis (Sec. S4 of the Supplementary Material), it significantly outperforms all other methods on the remaining metrics and shows substantial improvements over the baseline. Specifically, TTGA improves the baseline DSC by margins ranging from $1.1\%$ (on the specialized H-vmunet model for the skin lesion segmentation task) to $29.0\%$ (on the specialized HSNet model for the polyp segmentation task), and improves AUC from $1.7\%$ (on the specialized SegTran model for the optic cup and disc segmentation task in the fundus dataset) to $10.4\%$ (on the specialized HSNet model for the polyp segmentation task), and improves NSD from $0.7\%$ (on the specialized H-vmunet model for the skin lesion segmentation task) to $3.7\%$ (on the general-purpose nnU-Net-v2 model for the skin lesion segmentation task).

Regarding specific tasks, the segmentation of the optic disc/cup in the \textbf{fundus} dataset poses challenges such as boundary ambiguity and vascular occlusion. For nnU-Net-v2, due to the model's high intrinsic confidence, non-generative augmentation schemes demonstrate poor ability to assess the error regions. The null-text inversion scheme, being too close to the original image, also achieves low scores. For the specialized SegTran model, all comparative methods demonstrate high accuracy in error estimation for this task, with TTD slightly outperforming TTA. Images augmented by TTA do not alter the overlapping or interactive information between vessels and the optic disc/cup, whereas TTD, by modifying model inference details, yields marginally better results. In comparison, TTGA effectively addresses both boundary ambiguity and layout occlusion, achieving the best performance overall. 

In the task of \textbf{polyp} segmentation, lesion boundaries are relatively well-defined, and baseline methods exhibit low uncertainty estimation accuracy, indicating a high level of confidence in model predictions. Under such conditions, on the specialized HSNet model, TTD significantly outperforms TTA. Analysis reveals that simple input perturbations, as employed by TTA, exert limited influence on model inference. In contrast, TTD achieves superior performance by optimizing predictive tendencies through internal parameter perturbations. Although input-level perturbations contribute minimally in this context, TTGA still achieves competitive performance, ranking second to TTD with only a $0.3\%$ gap in DSC, while significantly outperforming other methods in terms of AUC and NSD. This suggests that TTGA’s modifications to input features have a more substantial impact on segmentation outcomes than conventional augmentation strategies. For the general-purpose nnU-Net-v2 model, the baseline estimation results are already quite accurate due to the relative clarity of the boundaries. Null-text inversion, being close to the original image, maintains these fine results. Other augmentation methods, however, actually degrade the error estimation, suggesting that excessive random augmentation can be detrimental when the baseline performance is strong. In contrast, our TTGA, which differentiates augmentation between foreground and background, surpasses all other methods.

For the \textbf{skin} segmentation task, which similarly involves significant boundary ambiguity but fewer occlusion issues. For the general-purpose nnU-Net-v2, contrary to the polyp task, the blurrier boundaries introduce more uncertainty. Here, the augmentation strength of methods other than DDIM inversion appears insufficient, leading to lower error estimation. Our TTGA, however, maintains augmentation robustness and exhibits the best performance. Furthermore, for DDIM inversion, despite its fine DSC performance, its NSD performance is the lowest. This indicates that its highly random augmentations produce poor boundaries compared to other methods, and it only excels in the overlap-based DSC metric. For the specialized H-vmunet model, TTA outperforms TTD. This outcome suggests that when the baseline uncertainty estimation is sufficiently accurate, certain error estimation methods may produce inferior results. Alterations to model parameters or input patterns could cause the model to deviate from the optimal configuration for the test data. Leveraging the capability of image generation models to produce realistic medical images, TTGA generates augmented data that aligns with the true image distribution while mitigating unforeseen and undesirable biases in the model or data, leading to superior performance. 

\subsection{On improving segmentation accuracy}
\begin{table*}[htbp]
\centering
\caption{Performance comparison of different methods on pixel-wise error estimation across three tasks, evaluated by the DSC AUC, HD95 between uncertainty estimation regions and segmentation error regions. The results are binarized at a threshold of 0.5 (for the 0-1 range) before contributing to the metric calculation. In each set of experimental results, the best-performing value is indicated in bold font, while the second-best is indicated underlined font.}\label{tab_acc}
\setlength{\tabcolsep}{3pt}
\begin{tabular}{lcccccccccc}
\toprule
\multirow{2}{*}{Method}  & \multicolumn{3}{c}{Fundus} & \multicolumn{3}{c}{Polyp} & \multicolumn{3}{c}{Skin} \\
\cmidrule(lr){2-4} \cmidrule(lr){5-7} \cmidrule(lr){8-10} 
& DSC $\uparrow$ & AUC $\uparrow$ & HD95 $\downarrow$ & DSC $\uparrow$ & AUC $\uparrow$ & HD95 $\downarrow$ & DSC $\uparrow$ & AUC $\uparrow$ & HD95 $\downarrow$ \\

\midrule
Baseline &  \multicolumn{9}{c}{nnU-Net-v2} \\
\cmidrule(lr){2-10}

Unaugmented & 91.1\scriptsize{$\pm$6.5} & 98.5\scriptsize{$\pm$1.7} & 16.9\scriptsize{$\pm$30.6} & 85.7{\scriptsize$\pm$22.5} & 96.8{\scriptsize$\pm$10.8} & 31.4{\scriptsize$\pm$70.8} & 86.6\scriptsize{$\pm$14.5} & 96.6\scriptsize{$\pm$6.7} & 41.8\scriptsize{$\pm$48.0} \\

DDIM Inv. & 90.9\scriptsize{$\pm$5.9} & \textbf{99.0\scriptsize{$\pm$1.9}} & \underline{16.5\scriptsize{$\pm$24.5}} & 74.6{\scriptsize$\pm$32.8} & 93.7{\scriptsize$\pm$16.6} & 54.2{\scriptsize$\pm$97.6} & 84.9\scriptsize{$\pm$17.4} & 96.4\scriptsize{$\pm$7.4} & 44.6\scriptsize{$\pm$56.9} \\

Null-text Inv. & 91.1\scriptsize{$\pm$6.6} & 98.5\scriptsize{$\pm$1.9} & 17.5\scriptsize{$\pm$32.6} & 85.8{\scriptsize$\pm$22.8} & 96.1{\scriptsize$\pm$11.8} & 34.1{\scriptsize$\pm$70.6} & 86.4\scriptsize{$\pm$14.6} & 96.2\scriptsize{$\pm$6.8} & 42.0\scriptsize{$\pm$48.2} \\

Deep Ensemble & \underline{91.2\scriptsize{$\pm$6.6} }& 98.2\scriptsize{$\pm$2.0} & 16.7\scriptsize{$\pm$30.5} & \underline{86.4{\scriptsize$\pm$18.7}} & 95.6{\scriptsize$\pm$9.9} & 30.2{\scriptsize$\pm$68.2} & \underline{86.7\scriptsize{$\pm$14.5}} & 94.1\scriptsize{$\pm$8.1} & 42.1\scriptsize{$\pm$48.6} \\

TTA & 91.1\scriptsize{$\pm$6.8} & 98.7\scriptsize{$\pm$1.7} & 16.9\scriptsize{$\pm$31.0} & 86.0{\scriptsize$\pm$22.4} & \underline{97.1{\scriptsize$\pm$10.6}} & \underline{30.0{\scriptsize$\pm$70.4}} & 86.6\scriptsize{$\pm$14.4} & \underline{97.3\scriptsize{$\pm$6.0}} & \underline{41.5\scriptsize{$\pm$47.7}} \\

\rowcolor[gray]{0.9} TTGA (\textbf{Ours}) & \textbf{91.3\scriptsize{$\pm$6.0}} & \textbf{99.0\scriptsize{$\pm$1.1}} & \textbf{16.1\scriptsize{$\pm$28.2}} & \textbf{86.7{\scriptsize$\pm$19.3}} & \textbf{97.8{\scriptsize$\pm$9.2}} & \textbf{28.4{\scriptsize$\pm$67.7}} & \textbf{86.9\scriptsize{$\pm$14.2}} & \textbf{97.9\scriptsize{$\pm$6.4}} & \textbf{41.4\scriptsize{$\pm$47.8}} \\

\midrule
Baseline & \multicolumn{3}{c}{SegTran} & \multicolumn{3}{c}{HSNet} & \multicolumn{3}{c}{H-vmunet} \\
\cmidrule(lr){2-4} \cmidrule(lr){5-7} \cmidrule(lr){8-10}

Unaugmented & 87.4\scriptsize{$\pm$10.5} & 99.6\scriptsize{$\pm$0.6} & 20.3\scriptsize{$\pm$13.2} & 84.5\scriptsize{$\pm$22.3} & 94.0\scriptsize{$\pm$10.8} & 48.9\scriptsize{$\pm$88.6} & \underline{85.2\scriptsize{$\pm$16.1}} & 97.4\scriptsize{$\pm$7.4} & 44.8\scriptsize{$\pm$52.1} \\

DDIM Inv. & 88.2\scriptsize{$\pm$8.2} & \underline{99.8\scriptsize{$\pm$0.2}} & 18.2\scriptsize{$\pm$11.2} & 69.7\scriptsize{$\pm$33.9} & 90.0\scriptsize{$\pm$17.0} & 70.0\scriptsize{$\pm$92.2} & 84.9\scriptsize{$\pm$17.4} & 97.1\scriptsize{$\pm$7.7} & 44.5\scriptsize{$\pm$56.9} \\

Null-text Inv. & 88.1\scriptsize{$\pm$10.1} & 99.7\scriptsize{$\pm$0.4} & 18.7\scriptsize{$\pm$12.6} & \underline{84.2\scriptsize{$\pm$23.8}} & 93.6\scriptsize{$\pm$11.8} & 52.4\scriptsize{$\pm$92.2} & 85.0\scriptsize{$\pm$16.2} & 97.5\scriptsize{$\pm$7.6} & 43.8\scriptsize{$\pm$52.3} \\

TTD & \underline{89.1\scriptsize{$\pm$9.2}} & 99.7\scriptsize{$\pm$0.4} & 17.2\scriptsize{$\pm$12.7} & 82.8\scriptsize{$\pm$23.5} & \underline{96.5\scriptsize{$\pm$8.7}} & 49.9\scriptsize{$\pm$86.8} & 84.7\scriptsize{$\pm$16.9} & 92.7\scriptsize{$\pm$9.3} & 44.6\scriptsize{$\pm$51.2} \\

TTA & 88.5\scriptsize{$\pm$10.7} & \underline{99.8\scriptsize{$\pm$0.4}} & \underline{16.3\scriptsize{$\pm$11.1}} & 84.1\scriptsize{$\pm$23.1} & 96.4\scriptsize{$\pm$8.7} & \underline{47.6\scriptsize{$\pm$86.6}} & 85.1\scriptsize{$\pm$16.8} & \underline{97.8\scriptsize{$\pm$6.8}} & \underline{42.7\scriptsize{$\pm$53.1}} \\

\rowcolor[gray]{0.9} TTGA (\textbf{Ours}) & \textbf{89.7\scriptsize{$\pm$8.2}} & \textbf{99.9\scriptsize{$\pm$0.2}} & \textbf{15.1\scriptsize{$\pm$9.6}} & \textbf{84.6\scriptsize{$\pm$22.0}} & \textbf{96.9\scriptsize{$\pm$8.4}} & \textbf{46.2\scriptsize{$\pm$83.8}} & \textbf{85.4\scriptsize{$\pm$16.9}} & \textbf{98.6\scriptsize{$\pm$6.5}} & \textbf{42.5\scriptsize{$\pm$51.5}} \\

\bottomrule
\end{tabular}
\end{table*}

This section investigates the effects of test-time perturbation strategies on segmentation accuracy. The comparative methodology aligns with that described in Sec.~\ref{tab_acc}, evaluating three primary approaches: TTD and Deep Ensemble, which introduce model perturbations, and TTA, which applies data perturbations. As a reference for generative strategies, we additionally report the results obtained using DDIM inversion and null-text Inversion. The baseline segmentation accuracy is established using the output of the segmentation model on unperturbed original images, with results summarized in Tab.~\ref{tab_acc}.

The proposed TTGA demonstrates superior performance across all DSC, AUC, and HD95 metrics on both the general-purpose and specialized models for all three tasks, reflecting stable and accurate alignment with the ground truth. The standard deviations of the metrics are also reported, with TTGA achieving the lowest values in most datasets, highlighting its robustness in improving segmentation outcomes. More specifically, TTGA improves the baseline DSC by margins ranging from $0.1\%$ (on the specialized HSNet model for the polyp segmentation task) to $2.3\%$ (on the optic cup and disc segmentation task in the fundus dataset). For AUC, the improvement ranges from $0.3\%$ (in the fundus segmentation task) to $2.9\%$ (on the specialized HSNet model for the polyp segmentation task). For HD95 (where lower is better), the reduction ranges from $0.4$ (on the general-purpose nnU-Net-v2 for the skin segmentation task) to $5.2$ (on the specialized SegTran model for the fundus segmentation task.

For \textbf{polyp} segmentation task, the results on the specialized HSNet model show that TTD exhibits the poorest overall performance and stability in polyp segmentation, with a substantial DSC decline relative to the baseline despite excelling in pixel-wise error estimation. This indicates that TTD introduces randomness that is less correlated with semantic content, potentially covering error regions but with uncertain effects on precise boundary delineation. Similarly, TTA yields a notable DSC decrease compared to the baseline, alongside a low AUC of baseline, highlighting the polyp segmentation model’s limited robustness to test image perturbations and potential generalization deficiencies. As an uncontrollable generative approach, DDIM inversion introduces strong randomness, leading to unpredictable effects on the segmentation results and causing significant accuracy degradation on certain datasets. In contrast, null-text inversion preserves most of the information from the original image, resulting in highly limited augmentation capability and thus producing outcomes that closely resemble the baseline. For the general-purpose nnU-Net-v2 on the same polyp task, the improvements from other data perturbation methods are limited; notably, the highly random DDIM inversion even causes a significant performance drop. This suggests that when a baseline model's performance is already high, data augmentation must be exceptionally precise to yield improvements. Interestingly, Deep Ensemble, which introduces model uncertainty, brings a large improvement, despite performing the worst in the error estimation task (Tab.~\ref{tab_error}). Conversely, our TTGA not only achieves the highest segmentation accuracy improvement but also secured the best error estimation performance. This demonstrates that our guided augmentation strategy, which combines semantic and identity information, is a uniquely robust solution.

For \textbf{fundus} and \textbf{skin} segmentation tasks, where target boundaries are less distinct than in polyp segmentation, test-time perturbation strategies exhibit greater potential to enhance accuracy. All methods surpass the baseline in both DSC and AUC. Specifically, model perturbation methods like Deep Ensemble and TTD excels in optic disc and cup segmentation within the fundus dataset, likely due to its ability to introduce model randomness that enhances inference variability in scenarios with foreground occlusion. Conversely, TTA on the specialized H-vmunet model for skin segmentation, where significant occlusions are absent, though TTD struggles to consistently control randomness magnitude, resulting in less stable outcomes for some samples. TTGA consistently outperforms both approaches across these diverse scenarios, demonstrating robust adaptability and superior performance.

The results underscore TTGA’s effectiveness in balancing segmentation accuracy and robustness across varied tasks and domains. While other methods offer task-specific advantages, their limitations in stability and generalization highlight TTGA’s potential as a versatile and reliable test-time perturbation strategy. These findings suggest that TTGA’s controlled augmentation approach mitigates the trade-offs observed in model uncertainty methods (like Deep Ensemble and TTD) and data uncertainty augmentation methods (like TTA), making it a promising method for improving segmentation performance in practical applications.

We observe in Tab.~\ref{tab_acc} that the segmentation accuracy gains for highly confident or potentially overfitted models, such as the specialized HSNet baseline, appear modest. This is a key finding that demonstrates our method's robustness. For these sensitive models, aggressive and uncontrolled augmentations are detrimental. For instance, the non-identity-preserving DDIM Inversion causes a catastrophic DSC drop from $84.5\%$ to $69.7\%$. TTD also degrades performance (to $82.8\%$). As visualized in Fig.~\ref{qualitative1}, these models often exhibit high confidence, and overly aggressive perturbations can lead to poor results. TTGA's strength lies in its "smarter," more conservative augmentation, which intelligently balances identity preservation with semantic randomness to avoid these destructive perturbations. Consequently, TTGA is the only perturbation-based method in this scenario to provide a stable, positive improvement (to $84.6$ DSC), highlighting its robust applicability even when other methods fail.

\section{Discussion}
\subsection{Sampling analysis}\label{Sampling_Analysis}
\begin{figure*}
\centering
\includegraphics[width=1.0\textwidth]{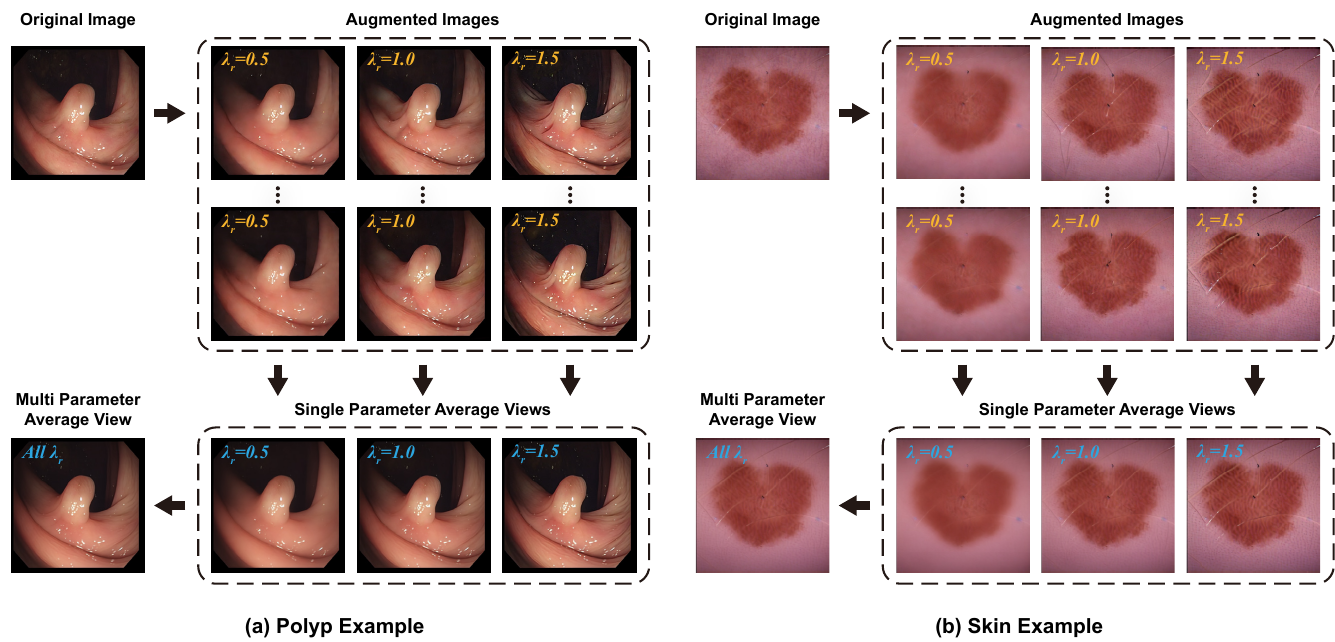}
\caption{Visualization of average views for two representative samples. For each identity guidance scale, augmented images exhibit local variations in detail. The average view of multiple augmentations closely resembles the original image, indicating that the augmentations are centered around the original image with minimal bias.} 
\label{fig_sampling}
\end{figure*}

In this subsection, we perform a qualitative analysis to examine the visual properties of images augmented using the proposed TTGA. Fig.~\ref{fig_sampling} showcases examples from two polyp segmentation and skin lesion segmentation where multiple augmented images are generated under varying identity guidance scales ($\lambda_r=0.5, 1.0, 1.5$). For each scale, two representative augmented images are presented, alongside their average views, to elucidate patterns introduced by the augmentation process.

Fig.~\ref{fig_sampling}(a) depicts a polyp case. Consistent with observations in Sec.~\ref{parameter_study}, the augmented images display distinct variations in the intensity of identity-related details across different $\lambda_r$. As the scale increases, identity features (such as structural details) grow more pronounced. Within a given scale, while the intensity of identity details remains consistent, localized differences in layout and texture emerge, reflecting TTGA’s controlled randomness. Crucially, essential features, such as the segmentation target’s overall contour, are faithfully preserved, whereas non-critical attributes, like background textures, exhibit diversified augmentation. This balance ensures that the augmented images retain semantic fidelity to the original while introducing valuable variability. To evaluate augmentation consistency, we computed the average view for each scale and across all scales. The average view at individual scales closely mirrors the original image in layout and contour, with differences primarily in the intensity of identity details. Notably, at a scale of $\lambda_r=1.0$ and when averaged across all scales, the average view exhibits striking similarity to the original image. This suggests that TTGA’s augmentation process is anchored around the original image, sampling variations that avoid introducing unexpected biases into segmentation outcomes. Thus, the augmented images are deemed reliable for downstream applications.

Fig.~\ref{fig_sampling}(b) illustrates a skin example, revealing trends analogous to those observed in the polyp case. Augmented images vary in detail intensity across scales, with localized randomness evident within the same scale. Unlike polyp images, however, skin lesion images feature less smooth textures and inherent noise. In the average view of these augmentations, key semantic attributes (such as color, morphology, and edge definitions) are well-maintained. Conversely, fine-grained textures appear smoothed, resembling a denoising effect. This indicates that TTGA is relatively insensitive to noise and excessive textural details, selectively augmenting non-critical features. Such augmentation may reduce distractions in segmentation tasks, enabling models to prioritize semantically significant information. The qualitative analysis highlights TTGA’s capability to produce augmented images that preserve essential semantic content while introducing controlled, task-relevant variability. This dual capacity ensures that the augmentations are both realistic and advantageous for enhancing segmentation robustness.

\subsection{Distribution analysis}
\begin{figure*}
\centering
\includegraphics[width=1.0\textwidth]{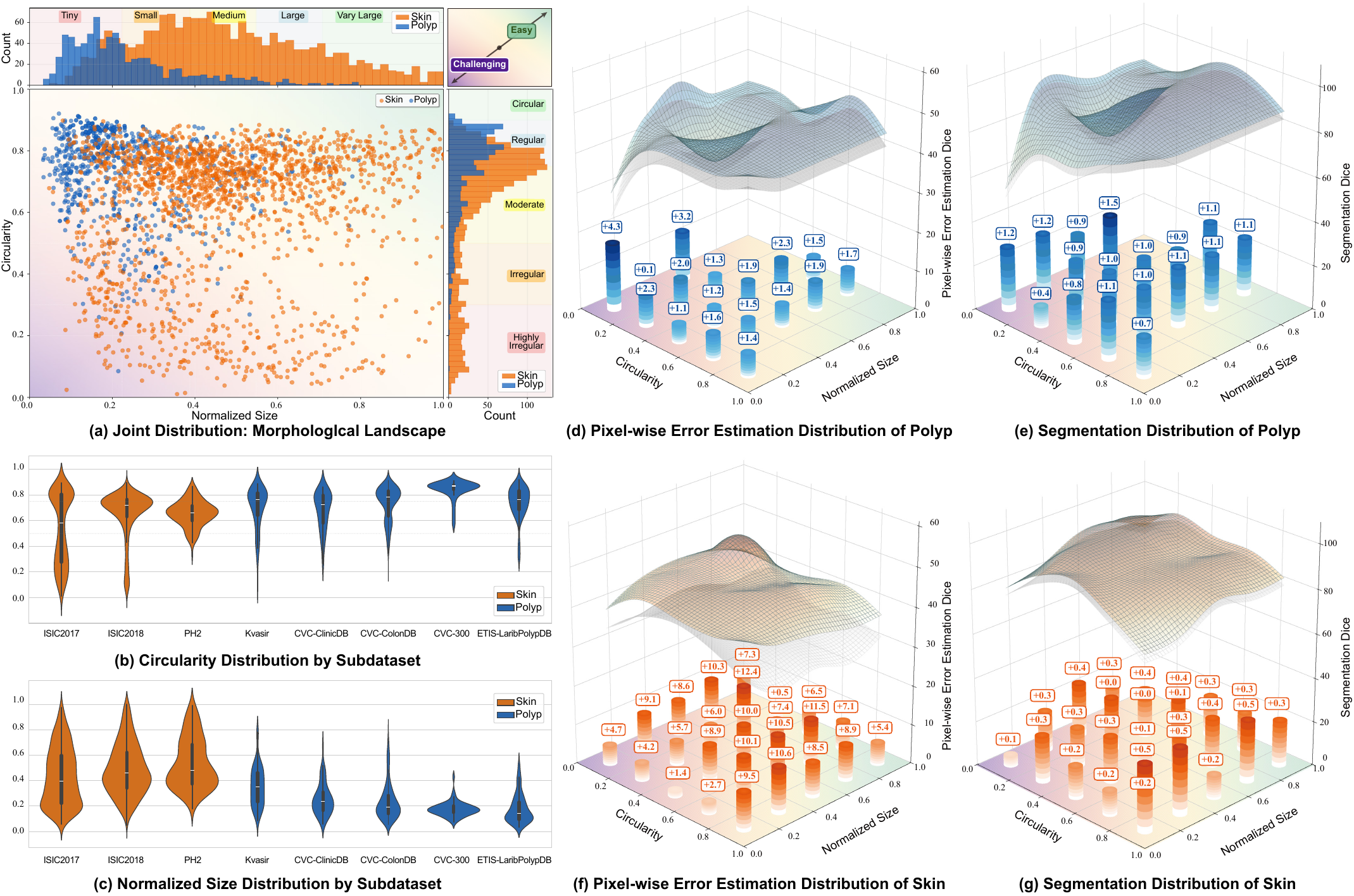}
\caption{Distribution Analysis of Morphological Landscape and Method Performance. (a) a joint scatter plot of Circularity versus Normalized Size for the Polyp (blue) and Skin (orange) datasets, with marginal histograms defining "Easy" versus "Challenging" morphological regions; (b) and (c) provide detailed violin plots of the individual Circularity and Normalized Size distributions, respectively, highlighting the distinct morphological characteristics across all datasets. The subsequent subplots (d-g) present 3D performance surfaces, comparing the baseline (gray wireframe}) to the consistently superior TTGA (blue wireframe for Polyp, orange wireframe for Skin) for both Pixel-wise Error Estimation Dice and Segmentation Dice. Note that for these metrics, higher values indicate better performance. The integrated bar charts quantify the absolute improvement provided by TTGA in each morphological bin, demonstrating its robust gains across all variations for both Polyp and Skin tasks.
\label{figure_distribution_analysis}
\end{figure*}

A potential limitation of segmentation methods is that their effectiveness may be confined to targets with simple shapes, which can affect generalizability. To investigate the influence of target morphology, we conducted a detailed distribution analysis, as presented in Fig.~\ref{figure_distribution_analysis}.

First, we visualized the "Morphological Landscape" of our two primary tasks, Skin and Polyp, by plotting the joint distribution of target Circularity and Normalized Size (Fig.~\ref{figure_distribution_analysis}(a)). We define Circularity as the metric $4\pi \times \text{Area} / \text{Perimeter}^2$, where a value of 1 indicates a perfect circle and values approaching 0 represent increasingly irregular or complex shapes. Normalized Size is defined as $\sqrt{\text{Area}_{\text{target}} / \text{Area}_{\text{image}}}$, ranging from 1 (large) down to 0 (small). The scatter plot and marginal histograms clearly reveal that these two datasets are morphologically distinct. The polyp dataset (blue) is highly concentrated in the "Circular" and "Regular" regions (Circularity > 0.8). Conversely, the skin dataset (orange) is broadly distributed, populating the "irregular" and "highly irregular" bins, which we defined as "challenging". This morphological divergence is further confirmed by the violin plots (Fig.~\ref{figure_distribution_analysis}(b,c)).

Having established this morphological diversity, we then analyzed the model's performance across this landscape. Fig.~\ref{figure_distribution_analysis}(d-g) present the 3D performance surfaces for the polyp and skin tasks. A core finding is immediately evident: the TTGA performance surface (solid) is systematically and globally higher than the Baseline performance surface (wireframe). This holds true for both pixel-wise error estimation Dice (d,f) and segmentation Dice (e,g).

This indicates that TTGA's advantage is not confined to a single morphological type. The bar charts in the figures quantify the absolute improvement within each bin, confirming that TTGA provides consistent positive gains across all bins. Crucially, Fig.~\ref{figure_distribution_analysis}(f,g) shows that for the skin task, TTGA delivers robust performance improvements even in the most "challenging" regions (low circularity, large size). This analysis confirms that the proposed method is not limited to "circular-ish" objects but effectively generalizes across the entire morphological landscape.

\subsection{Limitations and future works}
The proposed TTGA is a data-based perturbation approach designed for test-time perturbation. Across most scenarios, TTGA has demonstrated stable and reliable performance, yielding satisfactory results in uncertainty estimation and segmentation accuracy. However, its effectiveness in perturbation is limited when the segmentation model exhibits low sensitivity to data perturbations and a propensity for overfitting. As highlighted in the analysis of mixture solutions (Sec. S4 of the Supplementary Material), in the error estimation task for polyp segmentation, the standalone TTGA results were marginally inferior to those achieved by TTD. Nevertheless, integrating TTGA with model-based perturbation through a mixture solution led to substantial performance improvements. This suggests that incorporating model perturbation information can enhance TTGA’s capabilities. Given that TTGA is proposed as a general framework, this study explored only basic schemes and implemented a straightforward mixture solution with model perturbation in as detailed in Sec. S4 of the Supplementary Material. Future research could focus on incorporating model uncertainty guidance into the augmentation process. For example, expanding the denoising paths beyond the two outlined in Sec.\ref{sec_dual_path} to multiple paths could introduce additional control conditions, enabling the integration of model perturbation information into the augmentation process more effectively.

As a diffusion model-based generative method, TTGA offers varying levels of control over generative details depending on the sampling steps and intervals employed. For instance, a larger initial step introduces greater layout randomness, while different tissues or lesions may exhibit varying sensitivities to augmentation. To ensure simplicity and maintain fair comparisons, this study adopted fixed control parameters for most sampling processes. Future investigations could explore the influence of the sampling process on the control of generated image details more comprehensively. By designing task-specific, adaptive sampling processes that account for image-level semantic information (such as size, category, and boundary characteristics) TTGA could be extended to customized augmentation schemes. Such advancements could potentially improve its performance in a more targeted and effective manner.

Furthermore, regarding the ensemble strategy, this study employs a standard unweighted averaging approach to fuse the segmentation results from augmented samples. As discussed in Sec.~\ref{Sampling_Analysis}, since the generative augmentations are centered around the original image identity, averaging serves as a robust estimator to cancel out generative noise and restore the most stable prediction. However, we acknowledge that the estimated pixel-wise uncertainty map itself contains rich information that could be further utilized. Future work could explore uncertainty-guided fusion strategies, such as using the uncertainty map to assign spatial weights or dynamic thresholds during the ensemble process, potentially yielding even more precise boundary delineation.

\section{Conclusion}
This study introduces Test-Time Generative Augmentation (TTGA) as a significance advancement in the field of medical image segmentation, particularly for test-time enhancement. By harnessing a domain-adapted generative model, TTGA transcends the limitations of traditional test-time augmentation methods through its ability to generate multiple, semantically coherent views of test samples. The incorporation of region-specific augmentation, achieved via a novel masked null-text inversion technique and dual denoising paths, enables TTGA to improve both segmentation accuracy and uncertainty estimation at the pixel level. Extensive experimental validation across diverse medical imaging tasks—namely optic disc and cup segmentation, polyp segmentation, and skin lesion segmentation—demonstrates TTGA's superior performance over competing methods, especially under conditions of domain discrepancy. These results affirm TTGA's potential to mitigate challenges such as model overfitting and data variability, contributing to the development of more robust and generalizable segmentation systems. This work underscores the promise of generative augmentation as a cornerstone for advancing reliable medical image analysis.

\section*{Supplementary material}
Supplementary material associated with this article can be found, in the online version, at \url{http://dx.doi.org/10.1016/j.media.2025.103902}.

\section*{Acknowledgements}
This work was supported in part by the Major Research Plan of the National Natural Science Foundation of China under Grant (92370109), National Natural Science Foundation of China (62172223, 62201263, 62572245), Natural Science Foundation of Jiangsu Province (BK20252033, BK20220949), part by the ERC IMI (101005122), the H2020 (952172), the MRC (MC/PC/21013), the Royal Society (IEC/NSFC/211235), the NVIDIA Academic Hardware Grant Program, the SABER project supported by Boehringer Ingelheim Ltd, NIHR Imperial Biomedical Research Centre (RDA01), The Wellcome Leap Dynamic resilience program (co-funded by Temasek Trust), UKRI guarantee funding for Horizon Europe MSCA Postdoctoral Fellowships (EP/Z002206/1), UKRI MRC Research Grant, TFS Research Grants (MR/U506710/1), Swiss National Science Foundation (Grant No. 220785), and the UKRI Future Leaders Fellowship (MR/V023799/1, UKRI2738).

\end{document}